%% file: main.tex
\documentclass[10pt,twocolumn,letterpaper]{article}

\usepackage{wacv}              

\usepackage{graphicx}
\usepackage{amsmath}
\usepackage{amssymb}
\usepackage{booktabs}

\usepackage[]{adjustbox}
\usepackage{multirow}
\usepackage[normalem]{ulem}
\useunder{\uline}{\ul}{}
\usepackage{tikz}
\usepackage{xcolor}
\usepackage{url}

\usepackage[accsupp]{axessibility}  

\usepackage[percent]{overpic}
\definecolor{myred}{RGB}{211, 29, 29}
\definecolor{myblue}{RGB}{0, 171, 255}

\let\originalleft\left
\let\originalright\right
\renewcommand{\left}{\mathopen{}\mathclose\bgroup\originalleft}
\renewcommand{\right}{\aftergroup\egroup\originalright}

\newcommand{\deltaavg}{$<$$\delta^x_{avg}$}
\newcommand{\ft}[2]{{#1 \rightarrow #2}}   

\newcommand{\mat}[1]{{\mathbf{#1}}}
\newcommand{\vect}[1]{{\mathbf{#1}}}
\newcommand{\indexat}[2]{#1\left[#2\right]}
\newcommand{\sampleat}[2]{#1\left[#2\right]_s}
\newcommand{\stepfun}[1]{[\![#1]\!]}  
\newcommand{\bluedot}{{\color{myblue}\bullet}}
\newcommand{\ttimes}{\!\times\!}

\newcommand*\circled[1]{\tikz[baseline=(char.base)]{
            \node[shape=circle,draw,inner sep=0.8pt] (char) {#1};}}
\newcommand*\scircled[1]{{\scriptsize \circled{#1}}}

\newlength{\lewidth}
\usepackage{accsupp}

\usepackage[pagebackref,breaklinks,colorlinks]{hyperref}
\graphicspath{{figures/}}

\usepackage[capitalize]{cleveref}
\crefname{section}{Sec.}{Secs.}
\Crefname{section}{Section}{Sections}
\Crefname{table}{Table}{Tables}
\crefname{table}{Tab.}{Tabs.}

\def\wacvPaperID{1717} 
\def\confName{WACV}
\def\confYear{2024}

\begin{document}

\title{MFT: Long-Term Tracking of Every Pixel}

\author{Michal Neoral, Jon\'{a}\v{s} \v{S}er\'{y}ch, Ji\v{r}\'{i} Matas\\
  {\small CMP Visual Recognition Group, Faculty of Electrical Engineering, Czech Technical University in Prague} \\
  {\tt\small \{neoramic,serycjon,matas\}@fel.cvut.cz}
}
\maketitle

\begin{abstract}
We propose MFT -- Multi-Flow dense Tracker --  a novel method for dense, pixel-level, long-term tracking.
The approach exploits optical flows estimated not only between consecutive frames, but also for pairs of frames at logarithmically spaced intervals. It selects the most reliable sequence of flows  on the basis of estimates of its geometric accuracy and the probability of occlusion, both provided by a pre-trained CNN.
We show that MFT achieves competitive performance on the TAP-Vid benchmark, outperforming baselines by a significant margin, and tracking densely orders of magnitude faster than the state-of-the-art point-tracking methods.
The method is insensitive to medium-length occlusions and it is robustified by estimating flow with respect to the reference frame, which reduces drift.
\end{abstract}

\section{Introduction}
\label{sec:intro}

Reliable dense optical flow has a significant enabling potential for diverse computer vision applications, including  structure-from-motion, video editing, and augmented reality. 
Despite the widespread use of optical flow between consecutive frames for motion estimation in videos, generating consistent and dense long-range motion trajectories has been under-explored and remains a challenging task.

A simple baseline method for obtaining point-to-point correspondences in a video, e.g. for augmented reality, concatenates interpolated optical flow to form trajectories of a pixel, i.e. the set of projections of the pre-image of the pixel, for all frames in a sequence. 
However, such approach suffers from several problems: error accumulation leading to drift, sensitivity to occlusion and non-robustness, since a single poorly estimated optical flow damages the long-term correspondences for future frames.
This results in trajectories that quickly diverge and become inconsistent, particularly in complex scenes involving large motions, repetitive patterns and illumination changes.
Additionally, concatenated optical flow between consecutive frames cannot recover trajectories after occlusions.
Few optical flow approaches estimate occluded regions or uncertainty of estimated optical flow.

Another baseline approach --- matching every frame with the reference --- is neither prone to drift nor occlusions, but has other weaknesses. As the pose and illumination conditions change in the sequence, the matching problem becomes progressively more difficult. In the datasets used for evaluation in this paper, match-to-reference performs worse than consecutive frame optical flow concatenation. 

\begin{figure}
  \centering
  \input{figures/1d_chain}
  
  \caption{{\bf Overview of the MFT}. MFT tracks a query point ({\it black square}) by chaining optical flows.  Each chain consists of a previously computed chain from frame $0$ up to frame $(t-\Delta)$ ({\it dashed arrow, white dot}), and an optical flow vector computed between frames $(t-\Delta)$ and $t$ ({\it solid arrow}).
    MFT forms multiple candidate chains with varying $\Delta$.
    The best candidate ({\it black dot}) is selected according to uncertainty and occlusion scores.
    This is done in parallel, independently for each pixel in the reference frame.
  }
  \label{fig:chains}
\end{figure}
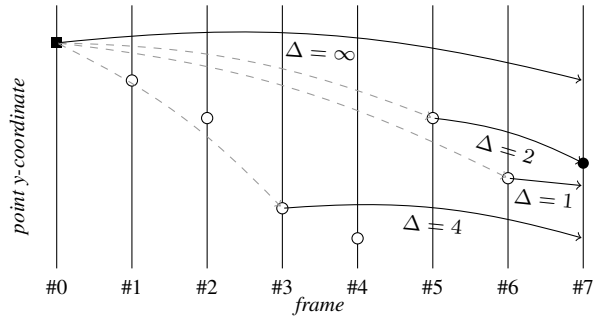

\begin{figure*}
\centering
\input{figures/video_editing}
\caption{{\bf MFT -- Multi-Flow Tracker application: video editing.}
A WOW! logo, inserted in frame $0$ of sequences from selected standard datasets~\cite{pont-tuset17davis,xu2018youtube}, propagated by MFT.
Frames at 0\%, 50\%, and 100\% of the sequence shown.
Full sequences in the supplementary.
}
\label{fig:mft_example}
\end{figure*}
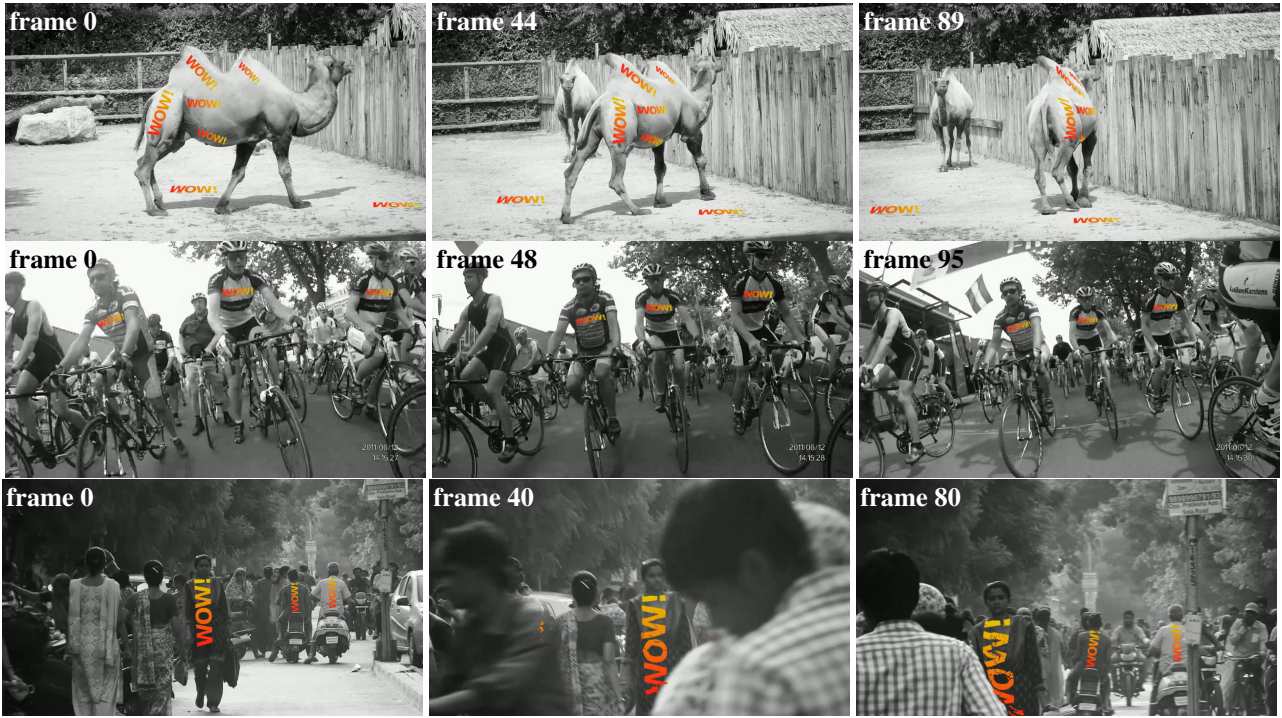

Addressing both weaknesses, we propose a novel method for dense long-term pixel-level tracking.
It is based on calculating flow not only for consecutive frames, but also for pairs of frames with logarithmically spaced time differences (see Fig.~\ref{fig:chains}).
We show that when equipped with suitable estimates of accuracy and of being occluded, a simple strategy for selecting the most reliable concatenation of the set of flows leads to dense and accurate long-term flow trajectories.
It is insensitive to medium-length occlusions and, helped by estimating the flow with respect to more distant frames, its drift is reduced.

The idea to obtain long-term correspondences by calculating a set of optical flows, rather than just flow between consecutive images, appeared for the first time in~\cite{crivelli2012optical}.
This led to a sequence of papers on the topic \cite{Crivelli2014, Conze2014, Conze2016}.
The performance of these early, pre-conv-net methods is difficult to assess.
They were mainly qualitatively, i.e. visually, tested on a few videos that are not available. 

The paper introduces the following contributions:
A point-tracking method that is (i) capable of tracking all pixels in a video based on CNN optical flow estimation, (ii) conceptually simple and can be trained and evaluated on a single customer grade GPU.
We show (iii) a simple yet effective strategy for selection of long-term optical flow chain candidates,
and (iv) how to select the most reliable candidate on the basis of spatial accuracy and occlusion probability obtained by small CNNs trained on synthetic data.
We publish the results and the method code \footnote{\url{https://github.com/serycjon/MFT}}.

  Experimentally the method outperforms baselines by a large margin and provide a good speed/performance balance, running orders of magnitude faster than the state-of-the-art for video point tracking~\cite{doersch2023tapir,karaev2023cotracker} when used for dense point tracking.
Fig.~\ref{fig:mft_example} shows an application of the proposed method for video editing.

\section{Related Work}
\label{sec:sota}

\paragraph{Object Tracking.}
Historically, object tracking algorithms~\cite{Bolme2010, Kalal2011, danelljan2018atom} estimated the location of an object specified in the first frame by a bounding box output in every frame of the video sequence. 
More recently~\cite{kristan2020eighth, perazzi2016davis}, the focus of tracking methods shifted to segmentation of the object or regions specified in the initial frame.
Nevertheless, algorithms that are model-free, i.e. are able to track any object specified in the first frame, do not provide point-to-point, dense correspondences. \vspace{-1em}

\paragraph{Structure-from-motion~(SfM) and SLAM}\hspace{-0.65em} are two related techniques that can be used for tracking points.
Although some methods can estimate the position of points densely~\cite{Gladkova2022}, they are limited to static scenes.
Non-rigid SfM techniques exist but are limited to a closed set of object categories since they require parametric 2D or 3D models~\cite{Li2020a, Biggs2020}.
N-NRSfM~\cite{Sidhu2020} is template-free, but prior to building a 3D model, it requires accurate 2D long-term tracks of points (chicken-egg problem).
Some approaches~\cite{Yang2021a, Yang2022a} utilize differentiable rendering or NeRF~\cite{mildenhall2021nerf} to create deformable 3D models for tracking points on surfaces.
However, their exhaustive computation makes them impractical for real-world usage. \vspace{-1em}

\paragraph{Optical flow}\hspace{-0.65em}estimation is a well-studied problem in computer vision that aims to estimate dense pixel-level displacements between consecutive frames~\cite{Horn1981}.
Modern methods employ deep learning techniques~\cite{Dosovitskiy2015, Ilg2017, Sun2018c, teed2020raft, Huang2022} trained on synthetic data.
State-of-the-art optical flow methods, such as RAFT~\cite{teed2020raft} and FlowFormer~\cite{Huang2022}, estimate optical flow from 4D correlation cost volume of features for all pixel-pairs.
While these methods achieve high accuracy for dense estimation of flow between pairs of consecutive frames,
estimating accurate flow between distant frames remains a problem, especially for large displacements or large object deformation.

Li \etal~\cite{Li2016a} combines feature matching and optical flow restricted with a deformable mesh.
NeuralMarker~\cite{Huang2022a} is trained to find correspondences between the template image and its distorted version inserted into random background image.
These approaches allow recovery from occluded regions. However, they are inapplicable for dense tracking in dynamic scenes with non-rigid objects.

To track points over multiple consecutive frames, some methods~\cite{Brox2010, Sundaram2010, Wang2013, Liu2017b} have proposed to concatenate estimated optical flow.
However, they cannot recover from partial occlusions.
Standard OF benchmarks~\cite{butler2012naturalistic,menze2015object} do not evaluate occlusion predictions and consequently most OF methods do not detect occlusions at all.
Moreover, concatenating optical flow results in error accumulation over time and induce drift in the tracked points.
Although some optical flow methods have been proposed to estimate the flow from more than two frames~\cite{Ren2018, Neoral2019}, they still operate in a frame-by-frame manner and do not handle partial occlusions well.
Therefore, achieving long-term, pixel-wise tracking with optical flow remains a challenging problem in computer vision. \vspace{-1em}

\paragraph{Multi-step-flow}\hspace{-0.65em}(MSF) algorithms~\cite{crivelli2012optical, Crivelli2012a, Crivelli2014} address the limitations of concatenation-based approaches for long-term dense point tracking.
These algorithms construct long-term dense point tracks by merging optical flow estimates computed over varying time steps.
This enables handling of temporarily occluded points by skipping them until they reappear.
However, they rely on the brightness constancy assumption, which leads to failure over distant frames.
The MSF approach has been updated in subsequent works~\cite{Conze2014, Conze2016} by introducing the multi-step integration and statistical selection (MISS) approach.
MISS generates a large number of motion path candidates by randomly selecting reference frames and weighting them based on estimated quality. The optimal candidate path is then determined through global spatial-smoothness optimization. However, these methods are computationally intensive and limited to tracking a small patch of a single object.

In comparison, our proposed MFT picks the best path based on occlusion and uncertainty estimated from correlation cost volume for individual optical flows.
Although some optical flow methods estimate occlusions~\cite{Ilg2018, Neoral2019, Hur2019, Zhao2020, Liu2021e, Zhang2022a} or uncertainty of estimated optical flow~\cite{Wannenwetsch2017, Ilg2018a, Yang2019}, state-of-the-art optical flow methods~\cite{teed2020raft, Huang2022} do not provide such estimates.
We are the first to employ estimation of occlusion and optical flow uncertainty for the dense and robust long-term tracking of points. \vspace{-1em}

\paragraph{Feature matching}\hspace{-0.65em} identifies corresponding points or regions between images or frames in a sequence.
Typically, feature matching is carried out sparsely on estimated keypoints.
While some dense estimation methods have been developed in the past~\cite{Brox2010}, they have not been able to match the performance of their sparse counterparts until recent advancements, such as the COTR approach~\cite{Jiang2021a}.
Note that feature matching is still performed only between pairs of frames and estimation of point positions may only be provided independently for target frames. \vspace{-1em}

\paragraph{Point tracking}\hspace{-0.65em}aims to track a set of physical points in a video as introduced in TAP-Vid~\cite{doersch2022tap}.
A baseline method TAP-Net~\cite{doersch2022tap} computes cost volume (similar to RAFT~\cite{teed2020raft}) for a single query point independently for each frame of the sequence.
A two-branch network then estimates the position and visibility of the query point in the targeted frame.
PIPs~\cite{harley2022particle} focuses on tracking points through occlusions by processing the video in fixed-sized temporal windows.
It does not re-detect the target after longer occlusions.
PIPs use test-time linking of estimated trajectories since it is limited to tracking in eight consecutive frames only.
Particle Video~\cite{sand2008particle} prunes tracked points on occlusion and creates new tracks on disocclusion, however these are not linked together.
TAPIR~\cite{doersch2023tapir} combines the per-frame point localization from TAP-Net~\cite{doersch2022tap} with a temporal processing inspired by PIPs~\cite{harley2022particle}, but uses a time-wise convolution instead of fixed size frame batches.
CoTracker~\cite{karaev2023cotracker} processes query points with a sliding-window transformer that enables multiple tracks to influence each other.
This works best when a single query point is tracked at a time, supported by an auxiliary grid of queries.
Compared to our proposed approach, these methods do not track densely, but instead focus on tracking individual query points.
OmniMotion~\cite{wang2023omnimotion} tracks densely.
It pre-processes the video by computing optical flow between all pairs of frames.
It represents the whole video with a quasi-3D volume, a NeRF\cite{mildenhall2021nerf}-like network and a set of 2D\(\leftrightarrow\)quasi-3D bijections.  The representation is globally optimized to obtain consistent motion estimates.

\begin{figure}
\centering
\input{figures/sintel/sintel_tikz}
\vspace{-12pt}
\caption{\textbf{Optical flow and occlusions.}
  OF methods are typically~\cite{teed2020raft,Huang2022} trained to ignore occlusions and to predict the ground-truth flow ({\it red}) even when occluded in the second frame.
  Continuing tracking after an occlusion would result in the target drifting to the occluding object.
  Example from Sintel~\cite{butler2012naturalistic}.
}
\label{fig:occlusion}
\end{figure}
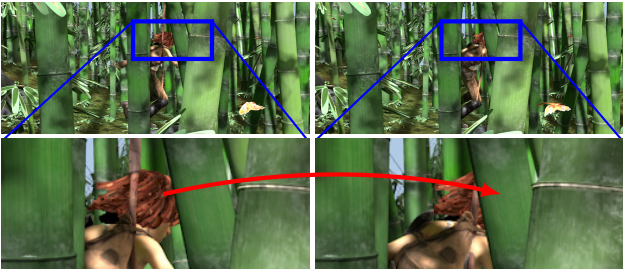

\begin{figure*}
  \centering
  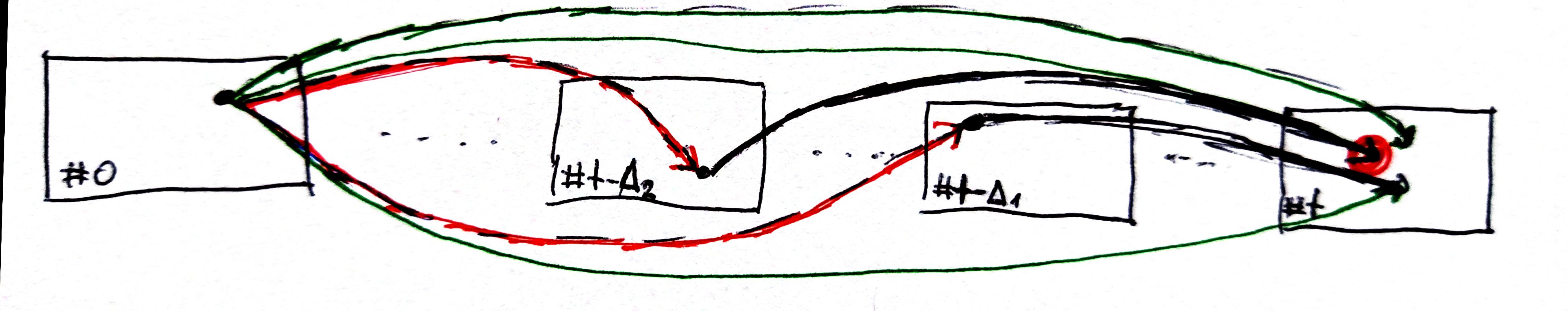
  \caption[dummy short caption]{{\bf Schematic explanation of the MFT tracking procedure.}
    At the current frame, time $t$ ({\it right}), the tracker creates a set of result candidates, each formed by a different chain of optical flows.
    In this example, the first candidate \scircled{3} is formed by chaining the result \scircled{1} previously computed in time $(t - \Delta_1)$ with flow \scircled{2} estimated between frames $(t - \Delta_1)$ and $t$.
    We use bilinear interpolation ({\it red}) to sample the flow field, since the positions in $(t - \Delta_1)$ usually do not align with the pixel grid.
    The flow \scircled{3} into the current frame $t$ is constructed by summing the two flow vectors.
    We repeat this procedure for $\Delta_2$, again summing the result \scircled{4} for frame $(t - \Delta_2)$ with \scircled{5} -- the bilinearly sampled flow field from $(t - \Delta_2)$ to $t$.
    When chaining the flows, we also chain their occlusion and uncertainty maps.
    Finally, we select the candidate (\scircled{3}, or \scircled{6}) with the lowest uncertainty score among the ones not occluded, or mark the result occluded when all candidates predict occlusion.
    Current point position shown in blue, grid-aligned flow vectors in black, interpolated flow vectors in red.
  }
  \label{fig:chaining}
\end{figure*}

\section{Method}
\label{sec:method}
The proposed method for long-term tracking of every pixel in a template is based on combining optical flow fields computed over different time spans, hence we call it Multi-Flow Tracker, or MFT in short.
Given a sequence of $H \times W$ -sized video frames $I_0, I_1, \dots, I_N$ and a list of positions on the reference (template) frame $\vect{p}_{i, 0} = (x_i, y_i), i \in \{1, \dots, HW\}$ the method predicts the corresponding positions $p_{i, t}$ in all the other frames $t \in \{1, \dots, N\}$, together with an occlusion flag $o_{i,t}$.
At time $t$, the MFT outputs are formed by combining the MFT result from a previous time $t-\Delta$, with the flow from $t-\Delta$ to the current frame $t$ (see Fig.~\ref{fig:chains}).
Note that this is not combining only two flows, but appending single flow to a previously computed, arbitrarily long chain of flows.
MFT constructs a set of candidate results with varying $\Delta$, then the best candidate is chosen independently for each template position.
To rank the candidates, MFT computes and propagates an occlusion map and an uncertainty map in addition to the optical flow fields.
Detecting occlusions is necessary to prevent drift to occluding objects as shown in Fig.~\ref{fig:occlusion}.
The position uncertainty serves to pick the most accurate of the candidates.
We now describe how the occlusion and uncertainty maps are formed, followed by a detailed description of the proposed MFT.

\subsection{Occlusion and Uncertainty}
Current optical flow methods typically compute the flow from a cost-volume inner representation and image features~\cite{teed2020raft,Huang2022,Sun2018c}.
Given a pair of input images, $I_a$ and $I_b$, the cost-volume encodes similarity between each position in $I_a$ and (possibly a subset of) positions in $I_b$.
We propose to re-use the cost-volume as an input to two small CNNs for occlusion and uncertainty estimation.
In both cases we use two convolutional layers with kernel size 3.
The first layer has 128 output channels and ReLU activation.
Both networks take the same input as the flow estimation head and each outputs a $H \times W$ map.

\noindent \textbf{Occlusion:} Similar to~\cite{Zhang2022a, Neoral2019, Hur2019}, we formulate the occlusion prediction as a binary classification.
The network should output 1 for any point in $I_a$ that is not visible in $I_b$ and 0 otherwise.
We train it on datasets with occlusion ground-truth labels (Sintel~\cite{butler2012naturalistic}, FlyingThings~\cite{Mayer2016}, and Kubric~\cite{greff2022kubric}) using standard cross-entropy loss.
The trained CNN achieves 0.96 accuracy on Sintel validation set.

\noindent \textbf{Uncertainty:}
We train the uncertainty CNN with the uncertainty loss function from~\cite{he2019bounding,bruggemann2023refign}
\begin{equation}
  \label{eq:unc-loss}
  \mathcal{L}_u = \frac{1}{2 \sigma^2}l_H(||\vec{x} - \vec{x}^*||_2) + \frac{1}{2}\log(\sigma^2)
\end{equation}
where $x$ is the predicted flow, $x^*$ the ground truth flow, $\sigma^2$ the predicted uncertainty and $l_H$ is the Huber loss function~\cite{huber1992robust}.
The uncertainty CNN predicts $\alpha = \log(\sigma^2)$ to improve numerical stability during training.
We output $\sigma^2$ during inference.

We sum the occlusion loss and
$\mathcal{L}_u$ weighted by $\frac{1}{5}$.
Note that we only train the occlusion and uncertainty networks, keeping the pre-trained optical flow fixed.

\subsection{MFT -- Multi-Flow Tracker}
\label{ssec:method_tracking}
The MFT tracker is initialized with the first frame of a video.
It then outputs a triplet $\overline{\mat{FOU}}_{\ft{0}{t}}=\left(\bar{\mat{F}}_\ft{0}{t}, \bar{\mat{O}}_{\ft{0}{t}}, \bar{\mat{U}}_{\ft{0}{t}}\right)$ at each consequent frame $I_t$.
The $\bar{\mat{F}}_{\ft{0}{t}}$ is a $H \times W \times 2$ map of position differences between frame number $0$ and $t$, in the classical optical flow format.
The $\bar{\mat{O}}_{\ft{0}{t}}$ and $\bar{\mat{U}}_{\ft{0}{t}}$ are $H \times W$ maps with the current occlusions and uncertainties respectively.
On the initialization frame, all three maps contain zeros only (no motion, no occlusion, no uncertainty),
on the first frame after initialization, the triplet is directly the output of the optical flow network and the proposed occlusion and uncertainty CNNs.
On all the following frames, the results are not the direct outputs of the network, but instead they are formed by chaining two $\left( \mat{F}, \mat{O}, \mat{U} \right)$ triplets together.

The MFT is parameterized by $D$, a set of time deltas.
We set $D = \{\infty, 1, 2, 4, 8, 16, 32\}$ (logarithmically spaced) by default.
For every $\Delta \in D$, we create a result candidate that is formed by chaining two parts -- a previously computed result $\overline{\mat{FOU}}_{\ft{0}{(t-\Delta)}}$ and a network output $\mat{FOU}_{\ft{(t-\Delta)}{t}}$ as shown in Fig.~\ref{fig:chaining}.
To keep the notation simple, we write $(t-\Delta)$, but in fact we compute $\max(0, t-\Delta)$ to avoid invalid negative frame numbers.

To do the chaining, we first define a new map $\bar{\mat{P}}_{(t-\Delta)}$ storing the point positions in time $(t-\Delta)$.
For each position $\vect{p} = (x, y)$ in the initial frame, the position in time $(t-\Delta)$ is calculated as
\begin{equation}
  \label{eq:P}
  \indexat{\bar{\mat{P}}_{(t-\Delta)}}{\vect{p}} = \vect{p} + \indexat{\bar{\mat{F}}_{\ft{0}{(t-\Delta)}}}{\vect{p}},
\end{equation}
where $\indexat{\mat{A}}{\vect{b}}$ means the value in a map $\mat{A}$ at integer spatial coordinates $\vect{b}$.
To form the candidate $\mat{F}_{\ft{0}{t}}^{\Delta}$, we add the optical flow $\mat{F}_{\ft{(t-\Delta)}{t}}$, sampled at the appropriate position to the motion between frames $0$ and $(t-\Delta)$.
\begin{equation}
  \label{eq:Fchain}
  \indexat{\mat{F}_{\ft{0}{t}}^{\Delta}}{\vect{p}} = \indexat{\bar{\mat{F}}_{\ft{0}{(t-\Delta)}}}{\vect{p}} + \sampleat{\mat{F}_{\ft{(t-\Delta)}{t}}}{\indexat{\bar{\mat{P}}_{(t-\Delta)}}{\vect{p}}}
\end{equation}
where $\sampleat{\mat{A}}{\vect{b}}$ means the value in a map $\mat{A}$ sampled at possibly non-integer spatial coordinates $\vect{b}$ with bilinear interpolation.
When chaining two occlusion scores, we take their maximum.
\begin{equation}
  \label{eq:Ochain}
  \indexat{\mat{O}_{\ft{0}{t}}^{\Delta}}{\vect{p}} = \max \left(\indexat{\bar{\mat{O}}_{\ft{0}{(t-\Delta)}}}{\vect{p}}; \sampleat{\mat{O}_{\ft{(t-\Delta)}{t}}}{\indexat{\bar{\mat{P}}_{(t-\Delta)}}{\vect{p}}} \right)
\end{equation}
Since we threshold the occlusion scores in the end to get a binary decision, this corresponds to an ``or'' operation -- the chain is declared occluded whenever at least one of its parts is occluded.

The uncertainties are chained by addition, as they represent the variance of the sum of flows, assuming independence of individual uncertainties.
\begin{equation}
  \label{eq:Uchain}
  \indexat{\mat{U}_{\ft{0}{t}}^{\Delta}}{\vect{p}} = \indexat{\bar{\mat{U}}_{\ft{0}{(t-\Delta)}}}{\vect{p}} + \sampleat{\mat{U}_{\ft{(t-\Delta)}{t}}}{\indexat{\bar{\mat{P}}_{(t-\Delta)}}{\vect{p}}}
\end{equation}

We repeat the chaining procedure for each $\Delta \in D$ to obtain up to $|D|$ different result candidates.
Finally, we select the best $\Delta$, $\Delta^{*}$ according to candidate uncertainty and occlusion maps.
In particular, we pick the $\Delta$ that has the lowest uncertainty score among the unoccluded candidates.
When all the candidates are occluded (occlusion score larger than a threshold $\theta_o$), all candidates are equally good and the first one is selected.
\begin{equation}
  \label{eq:BestDelta}
  \indexat{\mat{\Delta}^{*}}{\vect{p}} = \arg\!\min\limits_{\Delta \in D} \indexat{\mat{U}_{\ft{0}{t}}^\Delta}{\vect{p}} + \infty \cdot \stepfun{\indexat{\mat{O}_{\ft{0}{t}}^\Delta}{\vect{p}} > \theta_o},
\end{equation}
where $\stepfun{x}$ is the Iverson bracket (equal to $1$ when condition $x$ holds, $0$ otherwise).
Notice that we select the $\Delta^{*}$ independently for each position.
For example with $D = \{\infty, 1\}$, the flows are computed either directly between the template and the current frame ($\Delta = \infty$), or from the previous to the current frame ($\Delta = 1$) as in the traditional OF setup.
For some parts of the image, it is better to use $\Delta = \infty$, because having a direct link to the template does not introduce drift.
On the other hand, on some parts of the image the appearance might have significantly changed over the longer time span, making the direct flow not reliable at the current frame.
In such case a long chain of $\Delta = 1$ flows might be preferred.
Note that MFT usually switches back and forth between the used $\Delta$s during the tracking.
A single template query point might be tracked using a chain of $\Delta = 1$ flows for some time, then it might switch to the direct $\Delta = \infty$ flow for some frames (possibly undoing any accumulated drift), then back to $\Delta = 1$ and so on.

The final result at frame $t$ is formed by selecting the result from the candidate corresponding to $\Delta^{*}$ in each pixel, \eg, for the flow output $\bar{\mat{F}}_{\ft{0}{t}}$ we have
\begin{equation}
  \label{eq:PickingByDelta}
  \indexat{\bar{\mat{F}}_{\ft{0}{t}}}{\vect{p}} = \indexat{\mat{F}_{\ft{0}{t}}^{\indexat{\mat{\Delta}^{*}}{\vect{p}}}}{\vect{p}}
\end{equation}
Finally, MFT memorizes and outputs the resulting triplet $\overline{\mat{FOU}}_{\ft{0}{t}}$ and discard memorized results that will no longer be needed (more than $\max(D \setminus \{\infty\})$ frames old).
Given query positions $\vect{p}_{i, 0}$ on the template frame $0$, we compute their current positions and occlusion flags by bilinear interpolation of the $\overline{\mat{FOU}}$ result.
\begin{align}
  \label{eq:ConvertToTracklet}
  \vect{p}_{i, t} & = \vect{p}_{i, 0} + \sampleat{\bar{\mat{F}}_{\ft{0}{t}}}{\vect{p}_{i, 0}} \\
  o_{i, t}        & = \sampleat{\bar{\mat{O}}_{\ft{0}{t}}}{\vect{p}_{i, 0}}
\end{align}

\subsection{Implementation Details}
\label{sec:impl-details}
For the optical flow, we use the official RAFT~\cite{teed2020raft} implementation with author-provided weights.
Both the occlusion and the uncertainty CNNs operate on the same inputs as the RAFT flow regression CNN, \ie samples from the RAFT cost-volume, context features, and Conv-GRU outputs.
We train on Sintel~\cite{butler2012naturalistic}, FlyingThings~\cite{Mayer2016}, and Kubric~\cite{greff2022kubric}.
We sample training images with equal probability from each dataset.
Because the Kubric images are smaller than the RAFT training pipeline expects, we randomly upscale them with scale ranging between $3.2\ttimes$ and $4.6\ttimes$.
We train the occlusion and the uncertainty network for 50k iterations with the original RAFT training hyperparameters, which takes around 10 hours on a single GPU.

The MFT tracker is implemented in PyTorch and all the operations are performed on GPU.
Note that the optical flows and the occlusion and uncertainty maps can be pre-computed offline.
When the $\Delta = \infty$ is not included in $D$, the number of pre-computed flow fields needed to be stored in order to be able track forward or backward from any frame in a video is less than $N 2 |D|$.
Pre-computing flows for $\Delta = \infty$ (direct from template) and all possible template frames is not practical, as the number of stored flow fields grows quadratically with the number of frames $N$.
With the flows for other $\Delta$s pre-computed, MFT needs to compute just one OF per frame during inference, so the tracking speed stays reasonably fast.

On a GeForce RTX 2080 Ti GPU (i7-8700K CPU @ 3.70GHz), the chaining of the flow, occlusion and uncertainty maps takes approximately $1.3$ms for each $\Delta$ candidate with videos of $512 \times 512$ resolution.
On average, the preparation of all the result candidates takes $8$ms.
The per-pixel selection of the best one adds additional $0.6$ms.
Computing a single RAFT flow, including the extra occlusion and uncertainty outputs, takes $60$ms.
Altogether, the full MFT runs at 2.3FPS.
With pre-computed flows MFT runs at over 100FPS, making it suitable for interactive applications in, \eg, film post-production.
We set $\theta_o = 0.02$ empirically.

\begin{figure*}
  \setlength{\lewidth}{0.19\linewidth}
  \centering
  \raisebox{12mm}{\rotatebox{90}{Video}} %
  \begin{overpic}[width=\lewidth]{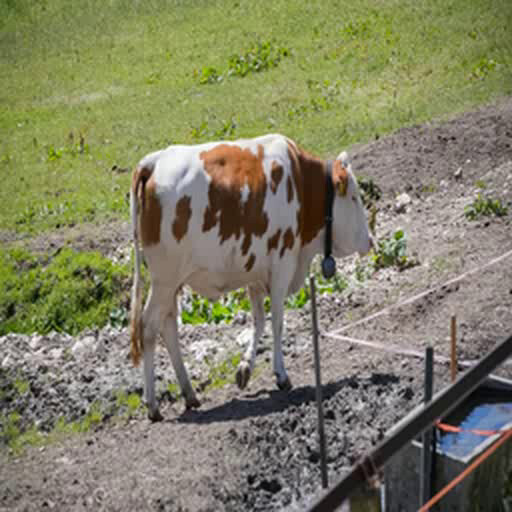}
     \put (2,90) {\color{white} \textbf{frame 0}}
  \end{overpic}
  \begin{overpic}[width=\lewidth]{checkerboards/input_video/00000026.jpg}
     \put (2,90) {\color{white} \textbf{frame 26}}
  \end{overpic}
  \begin{overpic}[width=\lewidth]{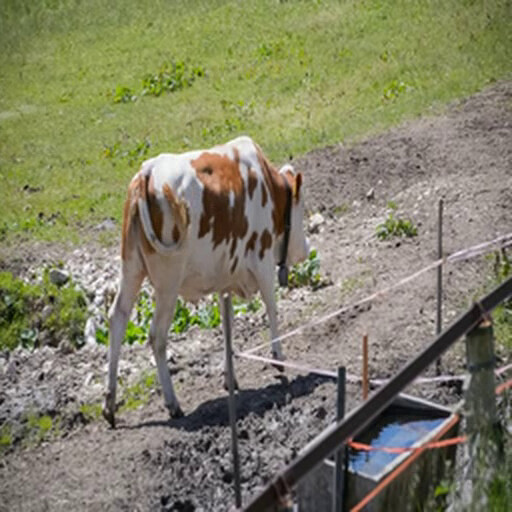}
     \put (2,90) {\color{white} \textbf{frame 52}}
  \end{overpic}
  \begin{overpic}[width=\lewidth]{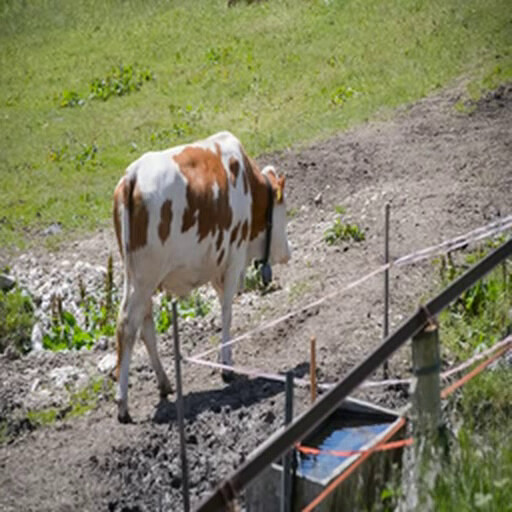}
     \put (2,90) {\color{white} \textbf{frame 77}}
  \end{overpic}
  \begin{overpic}[width=\lewidth]{checkerboards/input_video/00000103.jpg}
     \put (2,90) {\color{white} \textbf{frame 103}}
  \end{overpic}

  \raisebox{6mm}{\hspace{-0.6mm}\rotatebox{90}{Chain ($\Delta = 1$)}} %
  \includegraphics[width=\lewidth]{checkerboards/chain/00000000.jpg} %
  \includegraphics[width=\lewidth]{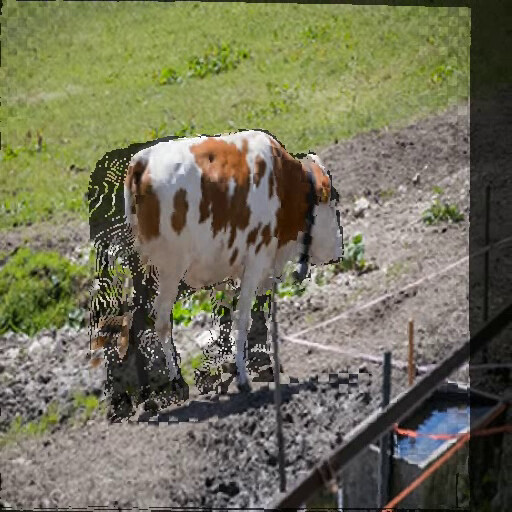} %
  \includegraphics[width=\lewidth]{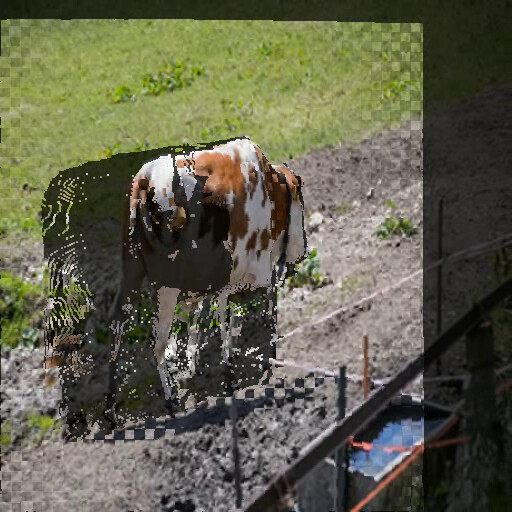} %
  \includegraphics[width=\lewidth]{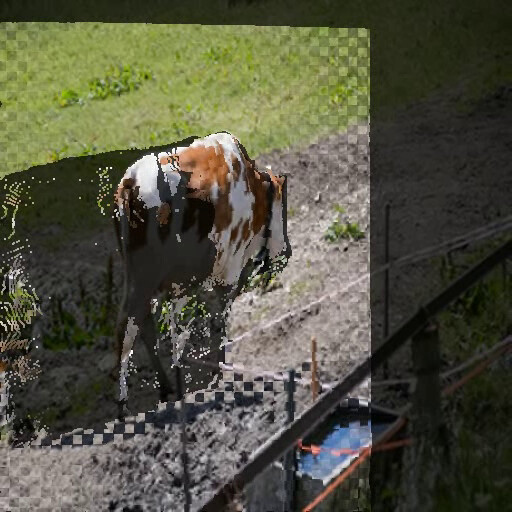} %
  \includegraphics[width=\lewidth]{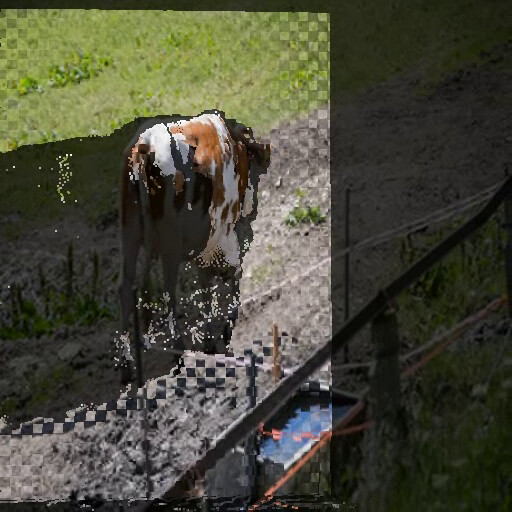}

  \raisebox{5mm}{\hspace{-0.6mm}\rotatebox{90}{Direct ($\Delta = \infty$)}} %
  \includegraphics[width=\lewidth]{checkerboards/direct/00000000.jpg} %
  \includegraphics[width=\lewidth]{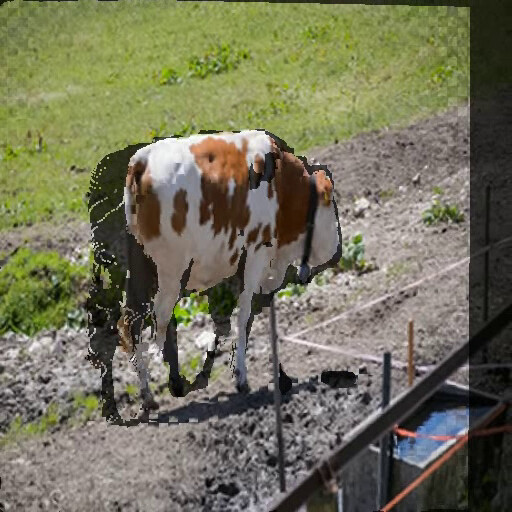} %
  \includegraphics[width=\lewidth]{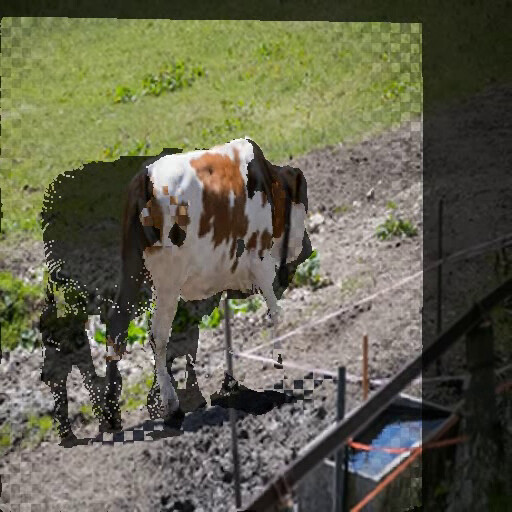} %
  \includegraphics[width=\lewidth]{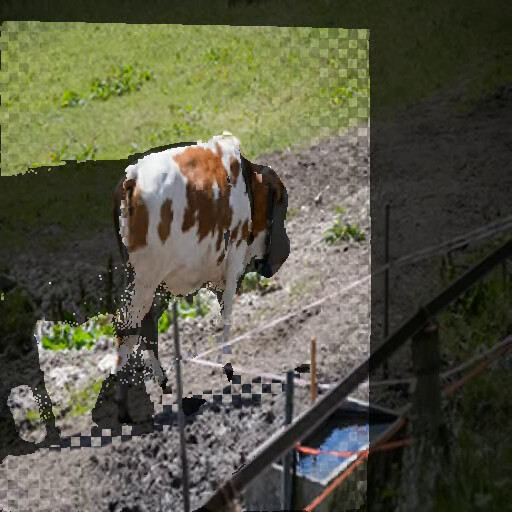} %
  \includegraphics[width=\lewidth]{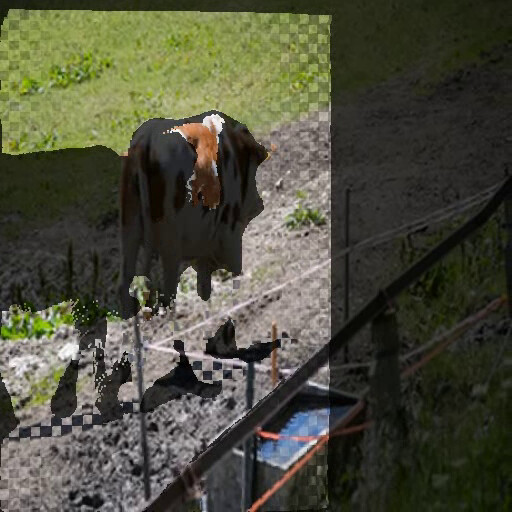}

  \raisebox{8mm}{\hspace{-0.6mm}\rotatebox{90}{\textbf{MFT} (ours)}} %
  \includegraphics[width=\lewidth]{checkerboards/MFT_occlthr002/00000000.jpg} %
  \includegraphics[width=\lewidth]{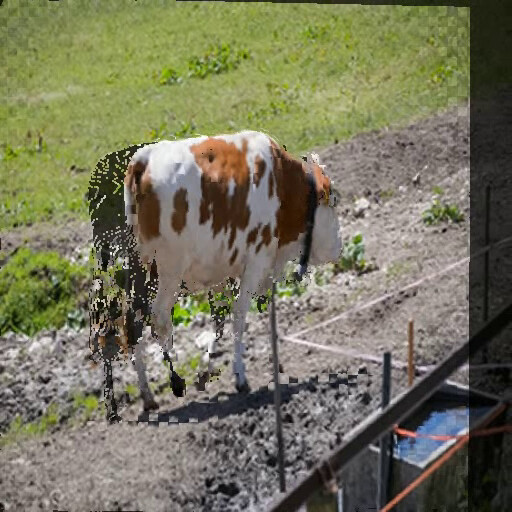} %
  \includegraphics[width=\lewidth]{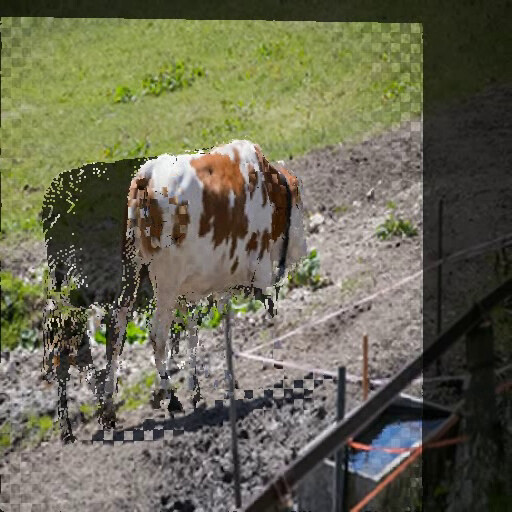} %
  \includegraphics[width=\lewidth]{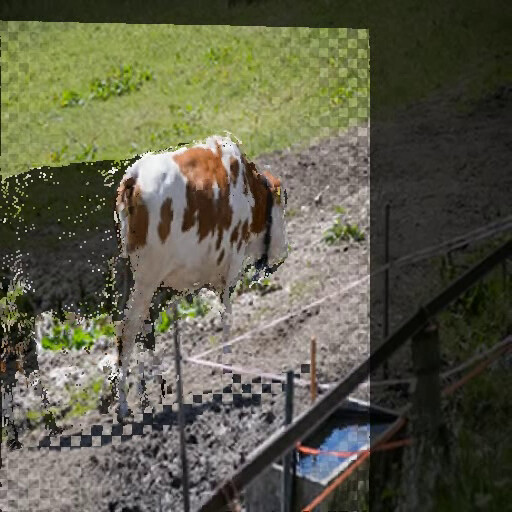} %
  \includegraphics[width=\lewidth]{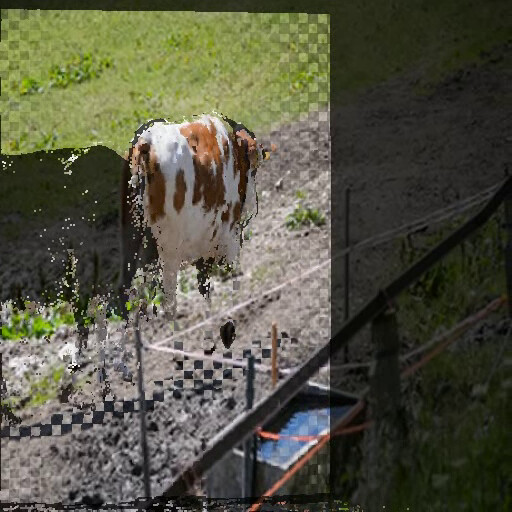}
  
  \caption{
    Result visualizations sampled at 25\%, 50\%, 75\% and 100\% of the input video ({\it top}) length.
    We take the first frame of a video and set its transparency with a checkerboard pattern.
    We then warp the resulting image using the outputs of each method and overlay the result on the current frame.
    The checkerboard pattern is visible when the tracking results are incorrect, or when the illumination changed between the template and the current frame.
    Pixels without a correspondence on the template frame are darkened.
    {\it Row 2:} simple flow chaining $\Delta=1$.  A short occlusion by the tail makes the tracker lose track in the back half of the cow.
    {\it Row 3:} direct flow $\Delta=\infty$.  The tracker survives the occlusion but loses track when the cow rotates away from the camera.
    {\it Bottom:} the proposed MFT handles both the short occlusion and the appearance change, tracking well on background and most of the cow's body.
    All trackers fail on the legs which are too thin for the RAFT optical flow.
    Best viewed zoomed-in on a screen.}
  \label{fig:checkerboard}
\end{figure*}

\section{Experiments}
\label{sec:experiments}
Since there is no benchmark for dense long-term point tracking, we evaluate the MFT on the recently introduced TAP-Vid DAVIS and TAP-Vid Kinetics datasets~\cite{doersch2022tap} for sparse point tracking.
The datasets consists of 30 videos from DAVIS 2017~\cite{pont-tuset17davis} and 1189 videos from Kinetics-700~\cite{carreira2017quo,carreira2019short} respectively, rescaled to $256 \times 256$ resolution, semi-automatically annotated with positions and occlusion flags of $\approx 20$ selected points.

\noindent \textbf{Evaluation protocol:}
The TAP-Vid benchmark uses two evaluation modes: ``first'' and ``strided''.
In the ``first'' mode, the tracker is initialized on the first frame where the currently evaluated ground-truth tracked point becomes visible, and is only evaluated on the following frames.
In the ``strided'' mode, the tracker is initialized on frames $0, 5, 10, \dots$ if the currently evaluated tracked point is visible in the given frame.
The tracker is then evaluated on both the following and the preceding frames, we thus run our MFT method two times, forward and backward in time, starting on the initialization frame.

\noindent \textbf{Evaluation metrics:}
The TAP-Vid benchmark uses three metrics.
The occlusion prediction quality is measured by occlusion classification accuracy (OA).
The accuracy of the predicted positions, \deltaavg, is measured by fraction of visible points with position error under a threshold, averaged over thresholds $1, 2, 4, 8, 16$.
Both occlusion and position accuracy are captured by Average Jaccard (AJ), see~\cite{doersch2022tap} for more details.

\subsection{Flow Delta Ablation}
\label{sec:flow-delta-ablation}
\begin{table}
\input{tables/tap_vid_ablation_method}
\caption{
\label{tab:tap_ablation_method} 
\textbf{TAP-Vid Davis benchmark -- evaluation of MFT on variants based on different sets } {$\mathbf{\mathit D}$} of time differences $\Delta$ used in optical flow; $\infty$ indicates OF between the template and the current frame.
Performance measured by occlusion accuracy (OA), position accuracy (\deltaavg), and combined measure~AJ. For definition of \deltaavg and AJ, see text.
Bold best, underline second.
}
\end{table}
In Table~\ref{tab:tap_ablation_method}, we show the impact of using different sets $D$ of $\Delta$s.
We evaluate two baselines -- (1) basic chaining of consecutive optical flows ($\Delta=1$), and (2), computing the optical flow directly between the template and the current frame ($\Delta=\infty$).
The first one performs better in all metrics, as the OF is computed on pairs of consecutive images, which it was trained to do,
and the test sequences are not long enough to induce significant drift by error accumulation.
Note that the performance in the strided evaluation mode is better, because the sequences are on average two times shorter and contain less occlusions.

Combining the basic chaining with the direct OF, line (3) in Table~\ref{tab:tap_ablation_method}, the performance increases in all metrics, showing the effectivity of the proposed candidate selection mechanism.
Row (4) is the full MFT method which achieves the overall best results.
The final experiment (5) works without the direct flow.
This means that we can pre-compute all the optical flows needed to track from any frame in any time direction, and store them in storage space proportional to the number of frames $2 N |D|$.
Note that attempting to do that with $\infty \in D$ would result in storage requirements proportional to $N^2$.
The last version achieves second best overall performance. 
Visual performance of the baselines and full MFT is shown in Fig.~\ref{fig:checkerboard}. 
All results in Table~\ref{tab:tap_ablation_method} were obtained on $2\times$ upscaled images as discussed in the next section which is equivalent to adding one upsampling layer to the RAFT feature pyramid.

\subsection{Input Resolution Ablation}
\label{sec:input-resolution-ablation}
\begin{table}
\input{tables/tap_vid_ablation_resolution}
\caption{
\label{tab:tap_ablation_resolution} 
\textbf{TAP-Vid Davis benchmark -- evaluation of MFT for different image resolutions.}
Performance measured as in Table~\ref{tab:tap_ablation_method}.
}
\end{table}
The official TAP-Vid benchmark is evaluated on videos rescaled to $256 \times 256$ resolution,
which is small compared with the RAFT training set.
Because of this, we upscale the $256 \times 256$ videos to $512 \times 512$ resolution.
In all the experiments, the output positions are scaled back to the $256 \times 256$ resolution for evaluation.
Rows (1) and (2) in Table~\ref{tab:tap_ablation_resolution} show that this upscaling improves the performance by a large margin on all three metrics.
This shows that RAFT is sensitive to input sizes, note that no information was added to the images when upscaling.

The aspect ratio of the original videos is changed during the scaling from full DAVIS resolution to the $256 \times 256$.
This makes the video contents appear distorted and changes the motion statistics.
Consequently we perform several experiments with varying video resolutions but keeping the original aspect ratio.
In the first two, (rows (3), (4) in Table~\ref{tab:tap_ablation_resolution}), we upsample the $256 \times 256$ videos.
This way we stick as close to the TAP-Vid protocol as possible, only requiring the original video aspect ratio as an extra input.
In (3), we keep the image height unchanged and only upscale the width such that the aspect ratio is not changed wrt the full resolution videos.
All the metrics improve compared to the no scaling variant (1).
Also, when we upscale the images to larger size (4), the performance increases.

In the last two rows (5), (6), we skip the TAP-Vid downscaling to $256 \times 256$ and instead downscale to the target resolution directly from the full-resolution DAVIS videos.
This preserves high-frequency details more than doing the downscale-upscale cycle.
Thanks to this, row (5) is better than (4), although the input resolution is the same in both.
Even larger resolution (6) again improves the \deltaavg and the AJ metric for the cost of small (below one percent point) decrease in occlusion accuracy.

Because we downscale directly from the full resolution, without the $256 \times 256$ intermediate step, the results of (5) and (6) are not directly comparable with the original TAP-Vid benchmark table, but are closer to a real-world scenario.

\subsection{Comparison With the State-of-the-Art}
\label{sec:sota-results}
On the TAP-Vid benchmark, the proposed MFT tracker performs third best, after the state-of-the-art sparse point-tracking methods~\cite{karaev2023cotracker,doersch2023tapir}, out-performing the other dense point-tracker OmniMotion~\cite{wang2023omnimotion}.
MFT runs at over 2FPS, which is orders of magnitude faster than the alternative methods evaluated densely, tracking every pixel and not just selected few.
The speed/performance balance makes MFT favorable for dense point-tracking.
Additionally, the optical flows can be pre-computed (only \(2 N \log N\) flows needed for a video of length $N$ with logarithmically spaced flow delta set $D$) resulting in tracking at over 100FPS from any frame in the video, both forward and backward.
This makes MFT a good candidate for interactive applications such as video editing.
The complete results, including the inference speeds, are shown in Table~\ref{tab:tap_vid_eval}.
Both MFT and OmniMotion~\cite{wang2023omnimotion} can be seen as post-processing of a set of RAFT optical flows.
The MFT strategy performs better than the complex model and global optimization in OmniMotion.

One MFT weakness we have observed are spurious re-detections.
MFT sometimes matches out-of-view parts of the template to visually similar parts of the current frame.
Single such incorrect re-detection can ``restart'' a flow chain, affecting the performance for the rest of the video.
A typical example is tracking of a point on a road surface.
When the camera moves such that the original point moves far out of view, the tracklet sometimes suddenly jumps to a newly uncovered patch of the road.
Both the appearance of the incorrectly matched point and its image context is often very similar to the template frame, \eg, a relatively texture-less black road some distance below a car wheel.

\noindent \textbf{BADJA evaluation.}  In addition to TAP-Vid DAVIS, we evaluate the MFT on BADJA~\cite{Biggs2018} benchmark with videos of animals annotated with 2D positions of selected joints.
The benchmark measures the percentage of points with position error under a permissive threshold $0.2 \sqrt{A}$, where $A$ is the area of the animal segmentation mask.
Thanks to this, the MFT performs well even though the ground truth points (joints) are located under the surface, and thus, MFT cannot track them directly.
In Table~\ref{tab:badja_eval}, we evaluate against the BADJA results of PIPs~\cite{harley2022particle} and their RAFT baseline.
In terms of median of the per-sequence results, MFT performs the best.
The mean score is affected by a single failure sequence, \emph{dog-a}, on which the dog turns shortly after the first frame, making most of the tracklets occluded.
The assumption that a joint can be approximately tracked by tracking a nearby point on the surface becomes invalid in such case.

\begin{table}
\input{tables/tap_vid_eval}

\caption{
\label{tab:tap_vid_eval}
\textbf{Evaluation on TAP-Vid benchmark.
  MFT performs well while being orders of magnitude faster than other methods when evaluated densely.}
Performance measured as in Table~\ref{tab:tap_ablation_method}.
Results for other methods are from~\cite{doersch2022tap,karaev2023cotracker,wang2023omnimotion,doersch2023tapir}.
FPS:~speed of dense (every pixel) tracking on \(512 \times 512\) video in Frames Per Second.
Speeds marked with \(\dagger\) were extrapolated from timing info in~\cite{doersch2023tapir,wang2023omnimotion}, details in supplementary.
}
\end{table}

\begin{table}
\input{tables/badja_eval}
\caption{
\label{tab:badja_eval} 
\textbf{BADJA~\cite{Biggs2018} benchmark -- evaluation of MFT against PIPs~\cite{harley2022particle}.}
Performance measured by the PCK-T measure, \ie, the percentage of points with error under a threshold.
Bold best.
Results for PIPs and RAFT from~\cite{harley2022particle}.
The labeled individual sequences include (a) bear, (b) camel, (c) cows, (d) dogs-a, (e) dog, (f) horse-h, and (g) horse-l.}
\end{table}

\section{Conclusions}
\label{sec:conclusions}
We proposed MFT -- a novel method for long-term tracking of every pixel on the template frame.
Its novelties include an introduction of two small CNNs estimating occlusion and flow uncertainty maps that are highly effective in selecting accurate flow chains that exploit flow computed both  between consecutive and non-consecutive frames.
MFT performs well on the point-tracking benchmark TAP-Vid~\cite{doersch2022tap}, and enables tracking all template pixels densely much faster (2.4 FPS vs 0.04 FPS) than the state-of-the-art point-trackers.
With pre-computed flows, MFT tracks densely at over 100FPS, enabling real-time interactivity for applications such as video editing.
Flow fields needed for MFT can be pre-computed offline, with storage requirements growing log-linearly with the video sequence length.
We also evaluated MFT on BADJA dataset, showing competitive performance on animal joint tracking.
We also show that accuracy of the popular RAFT optical flow increases significantly with input image resolution, even when upscaling low-resolution images which does not provide additional details.
We will publish the MFT code and weights.
{\bf Acknowledgements.}
This work was supported by Toyota Motor Europe,
by the Grant Agency of the Czech Technical University in Prague, grant 
No.SGS23/173/OHK3/3T/13, and
by the Research Center for Informatics project CZ.02.1.01/0.0/0.0/16\_019/0000765 funded by OP VVV.
{\small
\bibliographystyle{ieee_fullname}
\bibliography{sjo-bib,jabref}
}
\end{document}

%% file: figures/1d_chain.tex
\begin{tikzpicture}
  \tikzset{dot/.style={fill=black,circle}}

  \node [font=\footnotesize, rotate=90] at (-0.5, 1.5) {{\it point y-coordinate}};
  \node [font=\footnotesize] at (3.5, -0.5) {{\it frame}};
  \foreach \x in {0,1,...,7}
  {
    \draw (\x,0) -- (\x,3.5);
    \node [font=\footnotesize] at (\x,-0.25){\#\x};
  }

  \node[inner sep=0pt] (0) at (0, 3) {};
  \node[inner sep=0pt] (1) at (1, 2.5) {};
  \node[inner sep=0pt] (2) at (2, 2) {};
  \node[inner sep=0pt] (3) at (3, 0.8) {};
  \node[inner sep=0pt] (4) at (4, 0.4) {};
  \node[inner sep=0pt] (5) at (5, 2) {};
  \node[inner sep=0pt] (6) at (6, 1.2) {};
  \node[inner sep=0pt] (7) at (7, 2) {};
  \node[inner sep=0pt] (7_direct) at (7, 2.5) {};

  \node[inner sep=0pt] (7_D1) at (7, 1.1) {};
  \node[inner sep=0pt] (7_D2) at (7, 1.4) {};
  \node[inner sep=0pt] (7_D4) at (7, 0.4) {};

  \draw (0) node [rectangle,fill, inner sep=2pt] {};
  \draw (1) node [circle, fill=white, draw=black, inner sep=1.5pt] {};
  \draw (2) node [circle, fill=white, draw=black, inner sep=1.5pt] {};
  \draw (3) node [circle, fill=white, draw=black, inner sep=1.5pt] {};
  \draw (4) node [circle, fill=white, draw=black, inner sep=1.5pt] {};
  \draw (5) node [circle, fill=white, draw=black, inner sep=1.5pt] {};
  \draw (6) node [circle, fill=white, draw=black, inner sep=1.5pt] {};
  \draw [->] (0) edge [draw=black!40,dashed,bend left=10] (3);
  \draw [->] (0) edge [draw=black!40,dashed,bend left=10] (5);
  \draw [->] (0) edge [draw=black!40,dashed,bend left=10] (6);
  
  \draw [->] (0) edge [bend left=10] node [below, sloped, font=\footnotesize] {$\Delta = \infty$} (7_direct);
  \draw [->] (6) edge node [below, sloped, font=\footnotesize] {$\Delta = 1$} (7_D1);
  \draw [->] (5) edge [bend left=10] node [below, sloped, font=\footnotesize] {$\Delta = 2$} (7_D2);
  \draw [->] (3) edge [bend left=10] node [below, sloped, font=\footnotesize] {$\Delta = 4$} (7_D4);
  \draw (7_D2) node [circle,fill, inner sep=1.5pt] {};
\end{tikzpicture}

%% file: figures/video_editing.tex
\setlength{\lewidth}{0.32\linewidth}
\begin{overpic}[width=\lewidth]{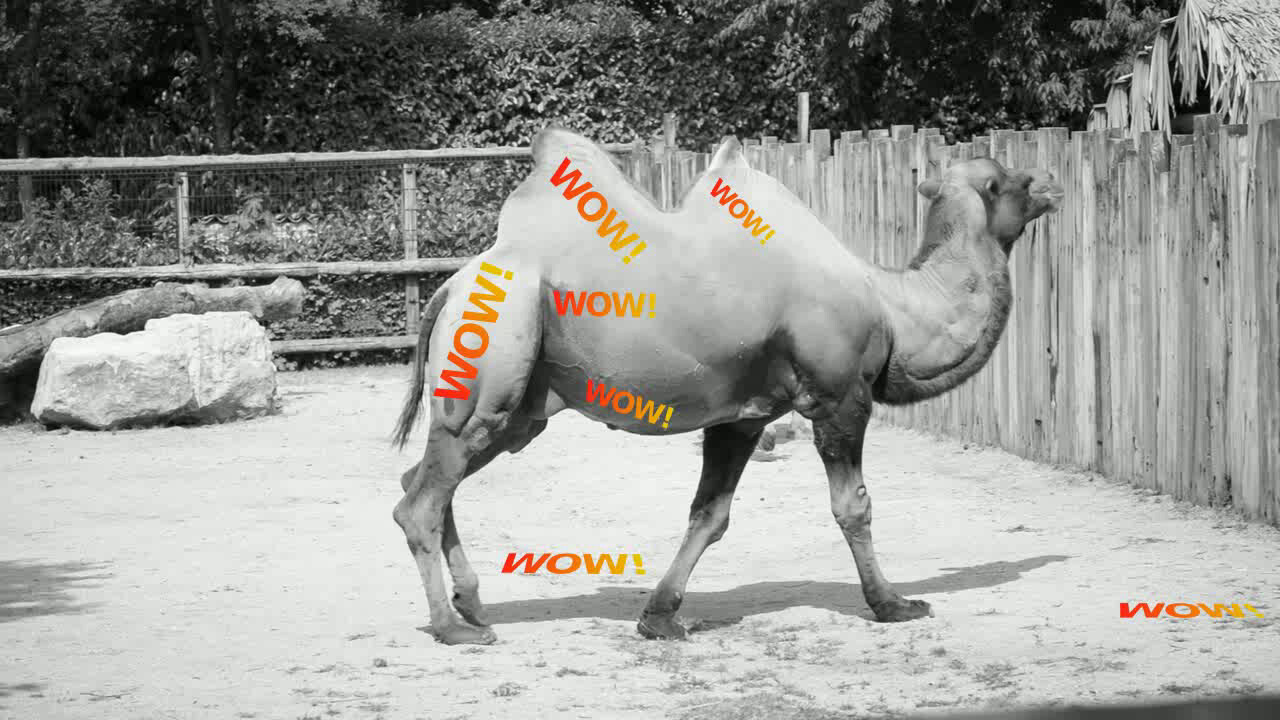}
     \put (1,50) {\color{white} \textbf{frame 0}}
\end{overpic}
\begin{overpic}[width=\lewidth]{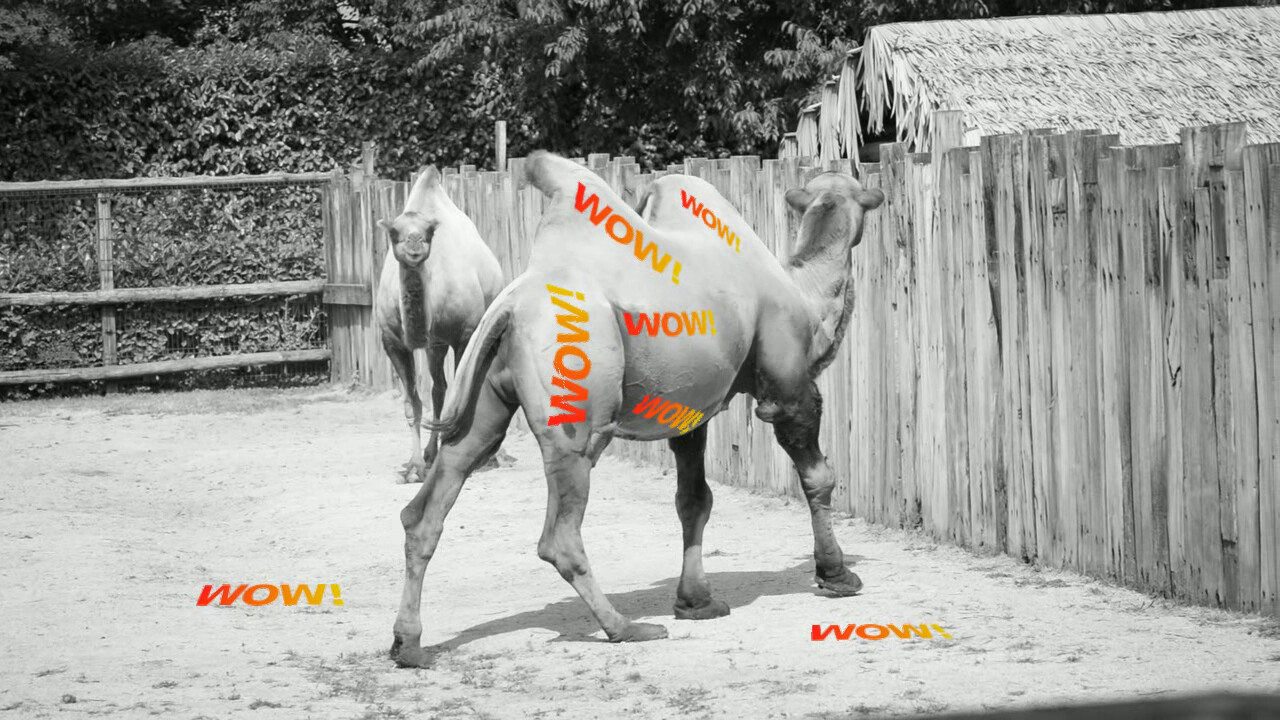}
     \put (1,50) {\color{white} \textbf{frame 44}}
\end{overpic}
\begin{overpic}[width=\lewidth]{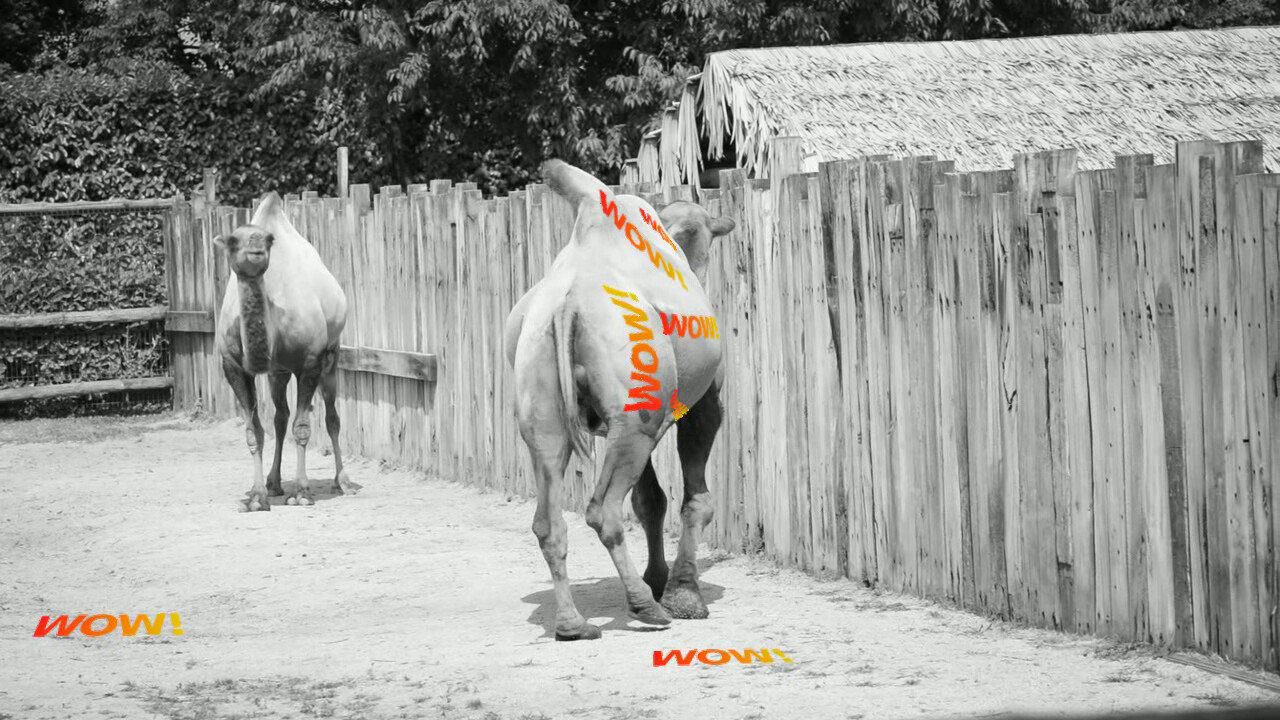}
     \put (1,50) {\color{white} \textbf{frame 89}}
\end{overpic}

\begin{overpic}[width=\lewidth]{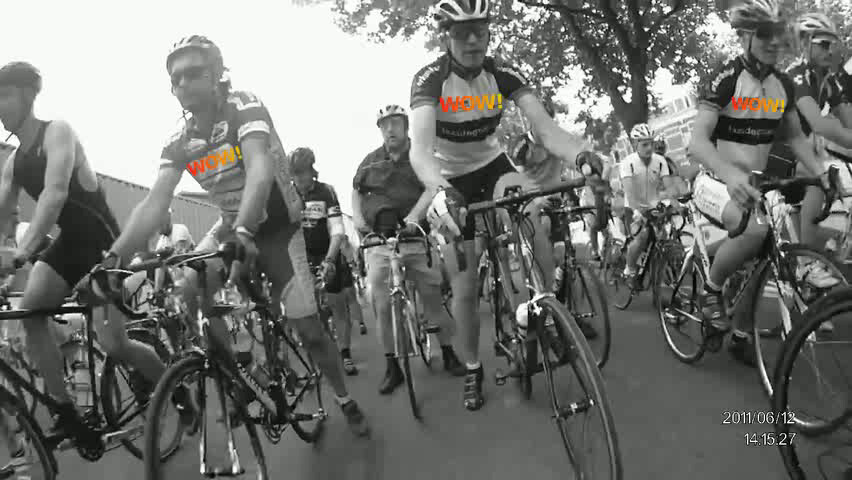}
     \put (1,50) {\color{black} \textbf{frame 0}}
\end{overpic}
\begin{overpic}[width=\lewidth]{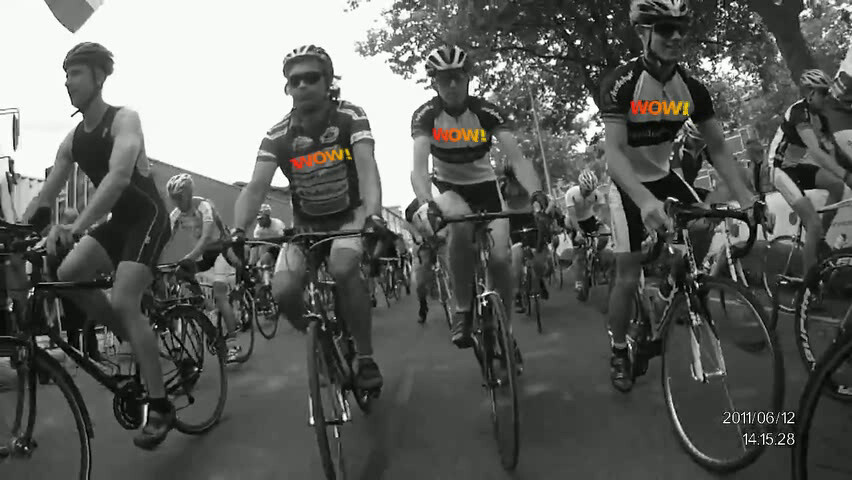}
     \put (1,50) {\color{black} \textbf{frame 48}}
\end{overpic}
\begin{overpic}[width=\lewidth]{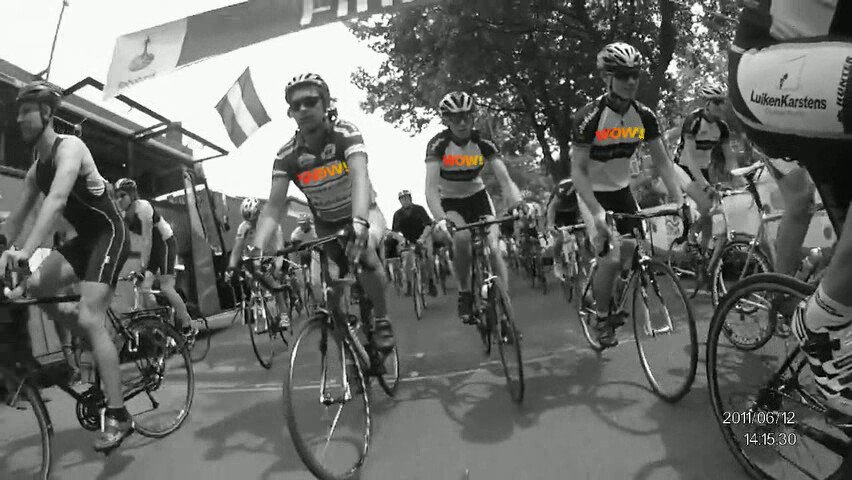}
     \put (1,50) {\color{black} \textbf{frame 95}}
\end{overpic}

\begin{overpic}[width=\lewidth]{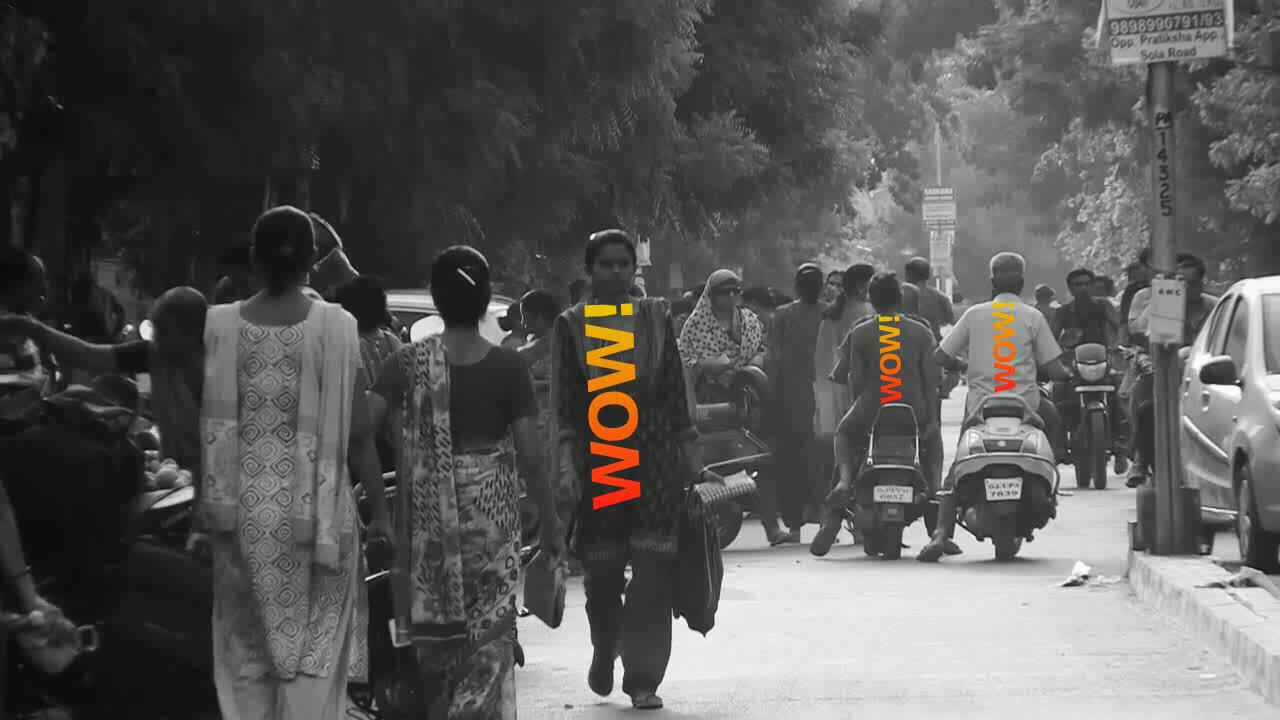}
     \put (1,50) {\color{white} \textbf{frame 0}}
\end{overpic}
\begin{overpic}[width=\lewidth]{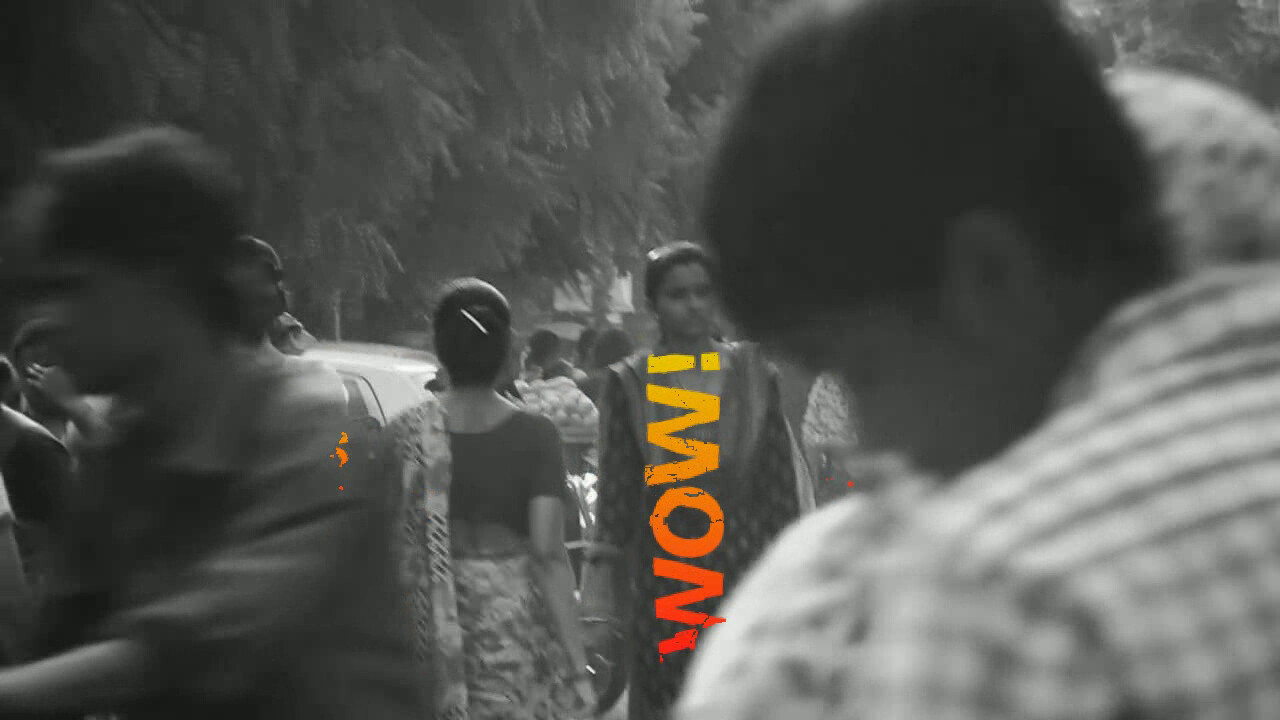}
     \put (1,50) {\color{white} \textbf{frame 40}}
\end{overpic}
\begin{overpic}[width=\lewidth]{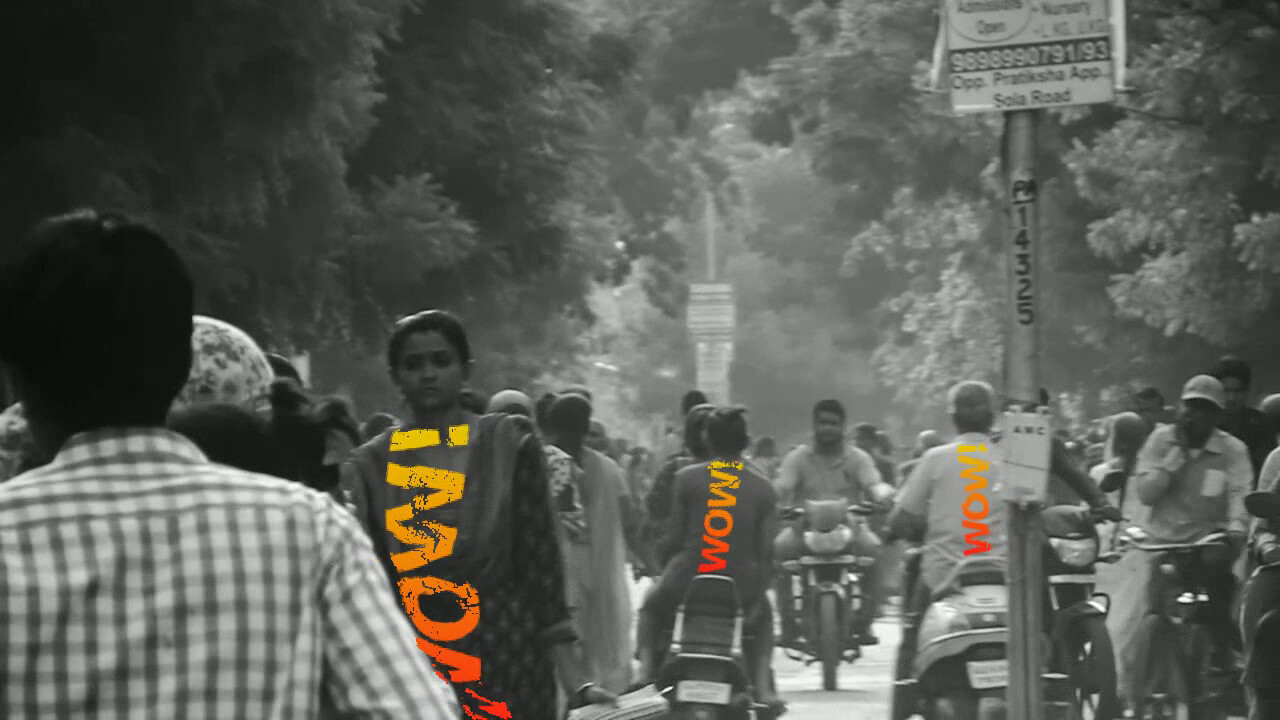}
     \put (1,50) {\color{white} \textbf{frame 80}}
\end{overpic}

%% file: figures/sintel/sintel_tikz.tex
\begin{tikzpicture}
   
    \node[anchor=south west,inner sep=0] at (0,1.8) {
        \includegraphics[width=0.49\linewidth]{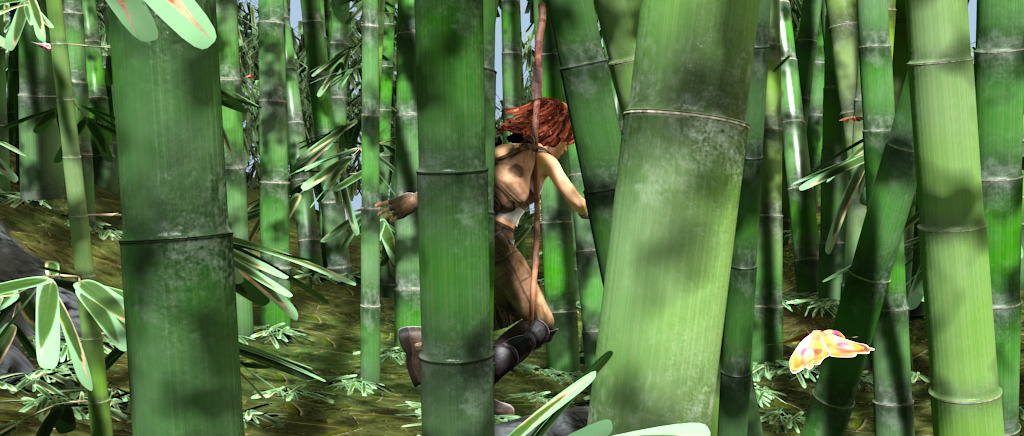}
        \includegraphics[width=0.49\linewidth]{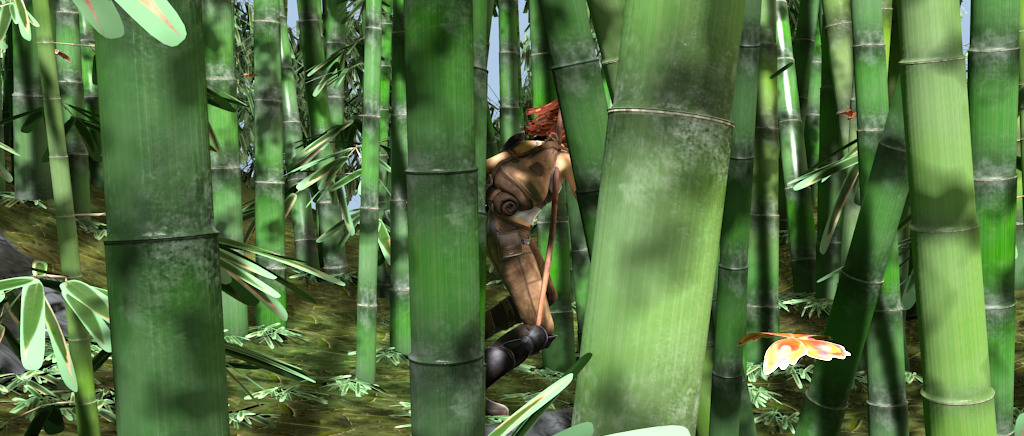}};
    
    \draw[blue,ultra thick] (2.8,3.3) rectangle (1.75,2.8);
    \draw[blue,ultra thick] (4.1+2.8,3.3) rectangle (4.1+1.75,2.8);
    
    \draw[blue,thick] (2.8,3.3) -- (4.07,1.75-0.03);
    \draw[blue,thick] (1.75,3.3) -- (0.02,1.75-0.03);
    \draw[blue,thick] (4.1+2.8,3.3) -- (4.17+4.08,1.75-0.03);
    \draw[blue,thick] (4.1+1.75,3.3) -- (4.18+0.00,1.75-0.03);

    \node[anchor=south west,inner sep=0] at (0,0) {
        \includegraphics[trim=15cm 9cm 12cm 2.5cm, clip=true, width=0.49\linewidth]{figures/sintel/frame_0030.jpg}
        \includegraphics[trim=15cm 9cm 12cm 2.5cm, clip=true, width=0.49\linewidth]{figures/sintel/frame_0031.jpg}
    };

    \draw[-latex,red,ultra thick] (2.15,1) arc (103:77:10cm);

\end{tikzpicture}

%% file: figures/chaining.pdf_tex
\begingroup%
  \makeatletter%
  \providecommand\color[2][]{%
    \errmessage{(Inkscape) Color is used for the text in Inkscape, but the package 'color.sty' is not loaded}%
    \renewcommand\color[2][]{}%
  }%
  \providecommand\transparent[1]{%
    \errmessage{(Inkscape) Transparency is used (non-zero) for the text in Inkscape, but the package 'transparent.sty' is not loaded}%
    \renewcommand\transparent[1]{}%
  }%
  \providecommand\rotatebox[2]{#2}%
  \newcommand*\fsize{\dimexpr\f@size pt\relax}%
  \newcommand*\lineheight[1]{\fontsize{\fsize}{#1\fsize}\selectfont}%
  \ifx\svgwidth\undefined%
    \setlength{\unitlength}{540.82310954bp}%
    \ifx\svgscale\undefined%
      \relax%
    \else%
      \setlength{\unitlength}{\unitlength * \real{\svgscale}}%
    \fi%
  \else%
    \setlength{\unitlength}{\svgwidth}%
  \fi%
  \global\let\svgwidth\undefined%
  \global\let\svgscale\undefined%
  \makeatother%
  \begin{picture}(1,0.13103388)%
    \lineheight{1}%
    \setlength\tabcolsep{0pt}%
    \put(0,0){\includegraphics[width=\unitlength,page=1]{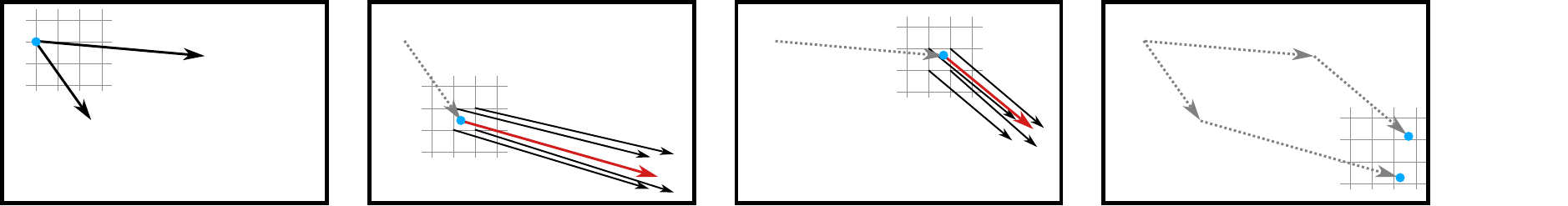}}%
    \put(0.18269977,0.11132579){\makebox(0,0)[lt]{\lineheight{1.25}\smash{\begin{tabular}[t]{l}\#$0$\end{tabular}}}}%
    \put(0.37407468,0.11132579){\makebox(0,0)[lt]{\lineheight{1.25}\smash{\begin{tabular}[t]{l}\#$t-\Delta_1$\end{tabular}}}}%
    \put(0.60705271,0.11132575){\makebox(0,0)[lt]{\lineheight{1.25}\smash{\begin{tabular}[t]{l}\#$t-\Delta_2$\end{tabular}}}}%
    \put(0.88440799,0.11132579){\makebox(0,0)[lt]{\lineheight{1.25}\smash{\begin{tabular}[t]{l}\#$t$\end{tabular}}}}%
    \put(0.08253049,0.08000312){\makebox(0,0)[lt]{\lineheight{1.25}\smash{\begin{tabular}[t]{l}\scircled{4} $\indexat{\bar{\mat{F}}_{\ft{0}{(t-\Delta_2)}}}{\bluedot}$\end{tabular}}}}%
    \put(0.24168335,0.00907249){\makebox(0,0)[lt]{\lineheight{1.25}\smash{\begin{tabular}[t]{l}{\color{myred}\scircled{2}} $\sampleat{\mat{F}_{\ft{(t-\Delta_1)}{t}}}{\bluedot}$\end{tabular}}}}%
    \put(0.49729741,0.04286733){\makebox(0,0)[lt]{\lineheight{1.25}\smash{\begin{tabular}[t]{l}{\color{myred}\scircled{5}} $\sampleat{\mat{F}_{\ft{(t-\Delta_2)}{t}}}{\bluedot}$\end{tabular}}}}%
    \put(0.73836625,0.02057834){\makebox(0,0)[lt]{\lineheight{1.25}\smash{\begin{tabular}[t]{l}$\scircled{3}=\scircled{1} + \scircled{2}$\end{tabular}}}}%
    \put(0.76648511,0.10565466){\makebox(0,0)[lt]{\lineheight{1.25}\smash{\begin{tabular}[t]{l}$\scircled{6}=\scircled{4}+\scircled{5}$\end{tabular}}}}%
    \put(0.01863218,0.03646559){\makebox(0,0)[lt]{\lineheight{1.25}\smash{\begin{tabular}[t]{l}\scircled{1} $\indexat{\bar{\mat{F}}_{\ft{0}{(t-\Delta_1)}}}{\bluedot}$\end{tabular}}}}%
  \end{picture}%
\endgroup%

%% file: tables/tap_vid_ablation_method.tex
\begin{center}
\begin{adjustbox}{max width=0.48\textwidth}
\begin{tabular}{lr|ccc|ccc}
\hline
             &                                    & \multicolumn{3}{c|}{DAVIS - first}                 & \multicolumn{3}{c}{DAVIS - strided}                \\
             & flow delta set $D$                 & AJ            & \deltaavg          & OA            & AJ            & \deltaavg          & OA            \\
\hline %
(1)          & $\{1\}$                            & 38.3          & 54.5               & 69.3          & 48.9          & 61.8               & 80.8          \\
(2)          & $\{\infty\}$                       & 38.3          & 50.8               & 65.5          & 47.9          & 58.0               & 76.3          \\
(3)          & $\{\infty, 1\}$                    & 46.4          & 63.7               & 76.7          & 55.0          & 68.1               & 85.8 \\
\textbf{(4)} & $\{\infty, 1, 2, 4, 8, 16, 32\}$   & {\ul 47.3}    & \textbf{66.8}      & \textbf{77.8} & \textbf{56.1} & \textbf{70.8}      & \textbf{86.9}  \\
(5)          & $\{1, 2, 4, 8, 16, 32\}$           & \textbf{47.4} & {\ul 66.2}         & {\ul 77.3}    & {\ul 55.7}    & {\ul 70.2}         & {\ul 86.5}    \\
\hline %
\end{tabular}
\end{adjustbox}
\end{center}

%% file: tables/tap_vid_ablation_resolution.tex
\begin{center}
\begin{adjustbox}{max width=0.48\textwidth}
\begin{tabular}{ll|ccc|ccc}
\hline
& \multirow{2}{*}{Resolution H$\times$W} & \multicolumn{3}{c|}{DAVIS - first}                              & \multicolumn{3}{c}{DAVIS - strided}     \\
&                                   & AJ                                 & \deltaavg & OA         & AJ         & \deltaavg & OA         \\
\hline %
(1) & 256$\times$256                     & 33.0                               & 47.7          & 70.2       & 41.4       & 54.6          & 83.6 \\
{\bf (2)} & 256$\times$256$\rightarrow$512$\times$512         & 47.3                               & 66.8          & 77.8       & 56.1       & 70.8          & 86.9       \\
\hline %
(3) & 256$\times$256$\rightarrow$256$\times$ratio       & 40.5                               & 58.5          & 76.9       & 49.2       & 63.8          & 86.4       \\
(4) & 256$\times$256$\rightarrow$480$\times$ratio       & 49.2                               & 69.2          & 77.9       & 58.8       & 73.9    & 87.7       \\
\hline %
(5) & orig res.$\rightarrow$480$\times$ratio       & {\ul 52.3}                                   & {\ul 71.9} & \textbf{79.5}                       &  {\ul 61.9}          & {\ul 76.1}              &  \textbf{88.8}         \\
(6) & orig res.$\rightarrow$720$\times$ratio       & \textbf{54.0}                      & \textbf{74.0} & {\ul 79.1} & \textbf{64.3}       & \textbf{78.7}          & {\ul 88.1}       \\
\hline %
\end{tabular}
\end{adjustbox}
\end{center} %
%
%

%% file: tables/tap_vid_eval.tex
\begin{center}
\begin{adjustbox}{max width=0.48\textwidth}
\begin{tabular}{ll|ccc|ccc|ccc}
  \hline
  \multirow{2}{*}{Method}                  & \multirow{2}{*}{FPS} & \multicolumn{3}{c|}{DAVIS - first} & \multicolumn{3}{c|}{DAVIS - strided} & \multicolumn{3}{c}{Kinetics - first}                     \\
                                           &                      & AJ                                 & \deltaavg                            & OA   & AJ   & \deltaavg & OA   & AJ   & \deltaavg & OA   \\ \hline
  TAP-Net~\cite{doersch2022tap}            & 0.11\(\dagger\)                 & 33.0                               & 48.6                                 & 78.8 & 38.4 & 53.1      & 82.3 & 38.4 & 54.4      & 80.6 \\
  PIPs~\cite{harley2022particle}           & 2e-4\(\dagger\)                 & -                                  & -                                    & -    & 42.0 & 59.4      & 82.1 & 31.7 & 53.7      & 72.9 \\
  OmniMotion~\cite{wang2023omnimotion}     & 2e-3\(\dagger\)                 & -                                  & -                                    & -    & 51.7 & 67.5      & 85.3 & -    & -         & -    \\ \hline
  \textbf{MFT (ours)}                      & {\bf 2.32}           & 47.3                               & 66.8                                 & 77.8 & 56.1 & 70.8      & 86.9 & 39.6 & 60.4      & 72.7 \\ \hline
  TAPIR~\cite{doersch2023tapir}            & 0.04\(\dagger\)                 & 56.2                               & 70.0                                 & 86.5 & 61.3 & 72.3      & 87.6 & 49.6 & 64.2      & 85.0 \\
  CoTracker~\cite{karaev2023cotracker}     & 0.04                 & 60.6                               & 75.4                                 & 89.3 & 64.8 & 79.1      & 88.7 & 48.7 & 64.3      & 86.5 \\
  \hline
\end{tabular}
\end{adjustbox}
\end{center}

%% file: tables/badja_eval.tex
\begin{center}
\begin{adjustbox}{max width=0.48\textwidth}
\begin{tabular}{l|ccccccc|cc}
\hline
 & a          & b         & c & d & e           & f & g       & Avg. & Med. \\ \hline
RAFT & 64.6          & 65.6          & 69.5 & 13.8  & 39.1          & 37.1    & 29.3          & 45.6 & 39.1 \\
PIPs & 76.3 & 81.6 & \textbf{83.2} & \textbf{34.2} & 44.0 & \textbf{57.4} & 59.5 & \textbf{62.3}  & 59.5\\
{\bf MFT}                           & \textbf{81.8} & \textbf{82.0} & 75.7 & 6.9   & \textbf{47.9} & 55.8    & \textbf{62.7} & 59.0 & \textbf{62.7} \\ \hline
\end{tabular}
\end{adjustbox}
\end{center}

%% file: main.bbl
\begin{thebibliography}{10}\itemsep=-1pt

\bibitem{Biggs2020}
Benjamin Biggs, Oliver Boyne, James Charles, Andrew Fitzgibbon, and Roberto
  Cipolla.
\newblock Who left the dogs out? 3d animal reconstruction with expectation
  maximization in the loop.
\newblock In {\em Computer Vision--ECCV 2020: 16th European Conference,
  Glasgow, UK, August 23--28, 2020, Proceedings, Part XI 16}, pages 195--211.
  Springer, 2020.

\bibitem{Biggs2018}
Benjamin Biggs, Thomas Roddick, Andrew Fitzgibbon, and Roberto Cipolla.
\newblock {C}reatures great and {SMAL}: {R}ecovering the shape and motion of
  animals from video.
\newblock In {\em ACCV}, 2018.

\bibitem{Bolme2010}
David~S Bolme, J~Ross Beveridge, Bruce~A Draper, and Yui~Man Lui.
\newblock Visual object tracking using adaptive correlation filters.
\newblock In {\em 2010 IEEE computer society conference on computer vision and
  pattern recognition}, pages 2544--2550. IEEE, 2010.

\bibitem{Brox2010}
Thomas Brox and Jitendra Malik.
\newblock Object segmentation by long term analysis of point trajectories.
\newblock In {\em European conference on computer vision}, pages 282--295.
  Springer, 2010.

\bibitem{bruggemann2023refign}
David Br{\"u}ggemann, Christos Sakaridis, Prune Truong, and Luc Van~Gool.
\newblock Refign: Align and refine for adaptation of semantic segmentation to
  adverse conditions.
\newblock In {\em Proceedings of the IEEE/CVF Winter Conference on Applications
  of Computer Vision}, pages 3174--3184, 2023.

\bibitem{butler2012naturalistic}
Daniel~J Butler, Jonas Wulff, Garrett~B Stanley, and Michael~J Black.
\newblock A naturalistic open source movie for optical flow evaluation.
\newblock In {\em Computer Vision--ECCV 2012: 12th European Conference on
  Computer Vision, Florence, Italy, October 7-13, 2012, Proceedings, Part VI
  12}, pages 611--625. Springer, 2012.

\bibitem{carreira2019short}
Joao Carreira, Eric Noland, Chloe Hillier, and Andrew Zisserman.
\newblock A short note on the kinetics-700 human action dataset.
\newblock {\em arXiv preprint arXiv:1907.06987}, 2019.

\bibitem{carreira2017quo}
Joao Carreira and Andrew Zisserman.
\newblock Quo vadis, action recognition? a new model and the kinetics dataset.
\newblock In {\em proceedings of the IEEE Conference on Computer Vision and
  Pattern Recognition}, pages 6299--6308, 2017.

\bibitem{Conze2014}
Pierre-Henri Conze, Philippe Robert, Tomas Crivelli, and Luce Morin.
\newblock Dense long-term motion estimation via statistical multi-step flow.
\newblock In {\em 2014 International Conference on Computer Vision Theory and
  Applications (VISAPP)}, volume~3, pages 545--554. IEEE, 2014.

\bibitem{Conze2016}
Pierre-Henri Conze, Philippe Robert, Tomas Crivelli, and Luce Morin.
\newblock Multi-reference combinatorial strategy towards longer long-term dense
  motion estimation.
\newblock {\em Computer Vision and Image Understanding}, 150:66--80, 2016.

\bibitem{Crivelli2012a}
Tom{\'a}s Crivelli, Pierre-Henri Conze, Philippe Robert, Matthieu Fradet, and
  Patrick P{\'e}rez.
\newblock Multi-step flow fusion: towards accurate and dense correspondences in
  long video shots.
\newblock In {\em British Machine Vision Conference}, 2012.

\bibitem{crivelli2012optical}
Tomas Crivelli, Pierre-Henri Conze, Philippe Robert, and Patrick P{\'e}rez.
\newblock From optical flow to dense long term correspondences.
\newblock In {\em 2012 19th IEEE International Conference on Image Processing},
  pages 61--64. IEEE, 2012.

\bibitem{Crivelli2014}
Tomas Crivelli, Matthieu Fradet, Pierre-Henri Conze, Philippe Robert, and
  Patrick P{\'e}rez.
\newblock Robust optical flow integration.
\newblock {\em IEEE Transactions on Image Processing}, 24(1):484--498, 2014.

\bibitem{danelljan2018atom}
Martin Danelljan, Goutam Bhat, Fahad~Shahbaz Khan, and Michael Felsberg.
\newblock Atom: Accurate tracking by overlap maximization.
\newblock In {\em Proceedings of the IEEE Conference on Computer Vision and
  Pattern Recognition}, pages 4660--4669, 2019.

\bibitem{doersch2022tap}
Carl Doersch, Ankush Gupta, Larisa Markeeva, Adria~Recasens Continente, Lucas
  Smaira, Yusuf Aytar, Joao Carreira, Andrew Zisserman, and Yi Yang.
\newblock {TAP-Vid}: A benchmark for tracking any point in a video.
\newblock {\em Advances in Neural Information Processing Systems}, 2022.

\bibitem{doersch2023tapir}
Carl Doersch, Yi Yang, Mel Vecerik, Dilara Gokay, Ankush Gupta, Yusuf Aytar,
  Joao Carreira, and Andrew Zisserman.
\newblock {TAPIR}: Tracking any point with per-frame initialization and
  temporal refinement.
\newblock In {\em Proceedings of the IEEE/CVF International Conference on
  Computer Vision (ICCV)}, pages 10061--10072, October 2023.

\bibitem{Dosovitskiy2015}
Alexey Dosovitskiy, Philipp Fischer, Eddy Ilg, Philip H{\"a}usser, Caner
  Haz{\i}rba{\c{s}}, Vladimir Golkov, Patrick van~der Smagt, Daniel Cremers,
  and Thomas Brox.
\newblock Flownet: Learning optical flow with convolutional networks.
\newblock In {\em ICCV}, pages 2758--2766, Dec. 2015.

\bibitem{Gladkova2022}
Mariia Gladkova, Nikita Korobov, Nikolaus Demmel, Aljo{\v{s}}a O{\v{s}}ep,
  Laura Leal-Taix{\'e}, and Daniel Cremers.
\newblock Directtracker: 3d multi-object tracking using direct image alignment
  and photometric bundle adjustment.
\newblock In {\em 2022 IEEE/RSJ International Conference on Intelligent Robots
  and Systems (IROS)}, pages 3777--3784. IEEE, 2022.

\bibitem{greff2022kubric}
Klaus Greff, Francois Belletti, Lucas Beyer, Carl Doersch, Yilun Du, Daniel
  Duckworth, David~J Fleet, Dan Gnanapragasam, Florian Golemo, Charles
  Herrmann, et~al.
\newblock Kubric: A scalable dataset generator.
\newblock In {\em Proceedings of the IEEE/CVF Conference on Computer Vision and
  Pattern Recognition}, pages 3749--3761, 2022.

\bibitem{harley2022particle}
Adam~W Harley, Zhaoyuan Fang, and Katerina Fragkiadaki.
\newblock Particle video revisited: Tracking through occlusions using point
  trajectories.
\newblock In {\em Computer Vision--ECCV 2022: 17th European Conference, Tel
  Aviv, Israel, October 23--27, 2022, Proceedings, Part XXII}, pages 59--75.
  Springer, 2022.

\bibitem{he2019bounding}
Yihui He, Chenchen Zhu, Jianren Wang, Marios Savvides, and Xiangyu Zhang.
\newblock Bounding box regression with uncertainty for accurate object
  detection.
\newblock In {\em Proceedings of the IEEE Conference on Computer Vision and
  Pattern Recognition}, pages 2888--2897, 2019.

\bibitem{Horn1981}
Berthold~KP Horn and Brian~G Schunck.
\newblock Determining optical flow.
\newblock {\em Artificial intelligence}, 17(1-3):185--203, 1981.

\bibitem{Huang2022a}
Zhaoyang Huang, Xiaokun Pan, Weihong Pan, Weikang Bian, Yan Xu, Ka~Chun Cheung,
  Guofeng Zhang, and Hongsheng Li.
\newblock Neuralmarker: A framework for learning general marker correspondence.
\newblock {\em ACM Transactions on Graphics (TOG)}, 41(6):1--10, 2022.

\bibitem{Huang2022}
Zhaoyang Huang, Xiaoyu Shi, Chao Zhang, Qiang Wang, Ka~Chun Cheung, Hongwei
  Qin, Jifeng Dai, and Hongsheng Li.
\newblock Flowformer: A transformer architecture for optical flow.
\newblock {\em Computer Vision--ECCV 2022: 17th European Conference, Tel Aviv,
  Israel, October 23--27, 2022, Proceedings, Part XVII}, pages 668--685, 2022.

\bibitem{huber1992robust}
Peter~J Huber.
\newblock Robust estimation of a location parameter.
\newblock {\em Breakthroughs in statistics: Methodology and distribution},
  pages 492--518, 1992.

\bibitem{Hur2019}
Junhwa Hur and Stefan; Roth.
\newblock Iterative residual refinement for joint optical flow and occlusion
  estimation.
\newblock In {\em Proceedings of the IEEE/CVF Conference on Computer Vision and
  Pattern Recognition}, pages 5754--5763, 2019.

\bibitem{Ilg2018a}
Eddy Ilg, Ozgun Cicek, Silvio Galesso, Aaron Klein, Osama Makansi, Frank
  Hutter, and Thomas Brox.
\newblock Uncertainty estimates and multi-hypotheses networks for optical flow.
\newblock In {\em Proceedings of the European Conference on Computer Vision
  (ECCV)}, pages 652--667, 2018.

\bibitem{Ilg2017}
Eddy Ilg, Nikolaus Mayer, Tonmoy Saikia, Margret Keuper, Alexey Dosovitskiy,
  and Thomas Brox.
\newblock Flownet 2.0: Evolution of optical flow estimation with deep networks.
\newblock {\em IEEE Conference on Computer Vision and Pattern Recognition
  (CVPR)}, 2017.

\bibitem{Ilg2018}
Eddy Ilg, Tonmoy Saikia, Margret Keuper, and Thomas Brox.
\newblock Occlusions, motion and depth boundaries with a generic network for
  disparity, optical flow or scene flow estimation.
\newblock In {\em Proceedings of the European conference on computer vision
  (ECCV)}, pages 614--630, 2018.

\bibitem{Jiang2021a}
Wei Jiang, Eduard Trulls, Jan Hosang, Andrea Tagliasacchi, and Kwang~Moo Yi.
\newblock Cotr: Correspondence transformer for matching across images.
\newblock In {\em Proceedings of the IEEE/CVF International Conference on
  Computer Vision}, pages 6207--6217, 2021.

\bibitem{Kalal2011}
Zdenek Kalal, Krystian Mikolajczyk, and Jiri Matas.
\newblock Tracking-learning-detection.
\newblock {\em IEEE transactions on pattern analysis and machine intelligence},
  34(7):1409--1422, 2011.

\bibitem{karaev2023cotracker}
Nikita Karaev, Ignacio Rocco, Benjamin Graham, Natalia Neverova, Andrea
  Vedaldi, and Christian Rupprecht.
\newblock {CoTracker}: It is better to track together.
\newblock {\em arXiv preprint arXiv:2307.07635}, 2023.

\bibitem{kristan2020eighth}
Matej Kristan, Ale{\v{s}} Leonardis, Ji{\v{r}}{\'\i} Matas, Michael Felsberg,
  Roman Pflugfelder, Joni-Kristian K{\"a}m{\"a}r{\"a}inen, Martin Danelljan,
  Luka~{\v{C}}ehovin Zajc, Alan Luke{\v{z}}i{\v{c}}, Ondrej Drbohlav, et~al.
\newblock The eighth visual object tracking vot2020 challenge results.
\newblock In {\em European Conference on Computer Vision}, pages 547--601.
  Springer, 2020.

\bibitem{Li2016a}
Wenbin Li, Darren Cosker, and Matthew Brown.
\newblock Drift robust non-rigid optical flow enhancement for long sequences.
\newblock {\em Journal of Intelligent \& Fuzzy Systems}, 31(5):2583--2595,
  2016.

\bibitem{Li2020a}
Xueting Li, Sifei Liu, Shalini De~Mello, Kihwan Kim, Xiaolong Wang, Ming-Hsuan
  Yang, and Jan Kautz.
\newblock Online adaptation for consistent mesh reconstruction in the wild.
\newblock {\em Advances in Neural Information Processing Systems},
  33:15009--15019, 2020.

\bibitem{Liu2021e}
Shuaicheng Liu, Kunming Luo, Nianjin Ye, Chuan Wang, Jue Wang, and Bing Zeng.
\newblock Oiflow: Occlusion-inpainting optical flow estimation by unsupervised
  learning.
\newblock {\em IEEE Transactions on Image Processing}, 30:6420--6433, 2021.

\bibitem{Liu2017b}
Yu Liu, Jianbing Shen, Wenguan Wang, Hanqiu Sun, and Ling Shao.
\newblock Better dense trajectories by motion in videos.
\newblock {\em IEEE transactions on cybernetics}, 49(1):159--170, 2017.

\bibitem{Mayer2016}
Nikolaus Mayer, Eddy Ilg, Philip Hausser, Philipp Fischer, Daniel Cremers,
  Alexey Dosovitskiy, and Thomas Brox.
\newblock A large dataset to train convolutional networks for disparity,
  optical flow, and scene flow estimation.
\newblock In {\em Proceedings of the IEEE conference on computer vision and
  pattern recognition}, pages 4040--4048, 2016.

\bibitem{menze2015object}
Moritz Menze and Andreas Geiger.
\newblock Object scene flow for autonomous vehicles.
\newblock In {\em Proceedings of the IEEE conference on computer vision and
  pattern recognition}, pages 3061--3070, 2015.

\bibitem{mildenhall2021nerf}
Ben Mildenhall, Pratul~P Srinivasan, Matthew Tancik, Jonathan~T Barron, Ravi
  Ramamoorthi, and Ren Ng.
\newblock {NeRF}: Representing scenes as neural radiance fields for view
  synthesis.
\newblock {\em Communications of the ACM}, 65(1):99--106, 2021.

\bibitem{Neoral2019}
Michal Neoral, Jan {\v{S}}ochman, and Ji{\v{r}}{\'i} Matas.
\newblock Continual occlusion and optical flow estimation.
\newblock In C.V. Jawahar, Hongdong Li, Greg Mori, and Konrad Schindler,
  editors, {\em Computer Vision -- ACCV 2018}, pages 159--174, Cham, 2019.
  Springer International Publishing.

\bibitem{perazzi2016davis}
F. Perazzi, J. Pont-Tuset, B. McWilliams, L.~Van Gool, M. Gross, and A.
  Sorkine-Hornung.
\newblock A benchmark dataset and evaluation methodology for video object
  segmentation.
\newblock In {\em Computer Vision and Pattern Recognition}, 2016.

\bibitem{pont-tuset17davis}
Jordi Pont-Tuset, Federico Perazzi, Sergi Caelles, Pablo Arbeláez, Alex
  Sorkine-Hornung, and Luc~Van Gool.
\newblock The 2017 davis challenge on video object segmentation.
\newblock {\em arXiv preprint arXiv:1704.00675v2}, 2017.

\bibitem{Ren2018}
Zhile Ren, Orazio Gallo, Deqing Sun, Ming-Hsuan Yang, Erik~B Sudderth, and Jan
  Kautz.
\newblock A fusion approach for multi-frame optical flow estimation.
\newblock In {\em 2019 IEEE Winter Conference on Applications of Computer
  Vision (WACV)}, pages 2077--2086. IEEE, 2019.

\bibitem{sand2008particle}
Peter Sand and Seth Teller.
\newblock Particle video: Long-range motion estimation using point
  trajectories.
\newblock {\em International Journal of Computer Vision}, 80:72--91, 2008.

\bibitem{Sidhu2020}
Vikramjit Sidhu, Edgar Tretschk, Vladislav Golyanik, Antonio Agudo, and
  Christian Theobalt.
\newblock Neural dense non-rigid structure from motion with latent space
  constraints.
\newblock In {\em Computer Vision--ECCV 2020: 16th European Conference,
  Glasgow, UK, August 23--28, 2020, Proceedings, Part XVI 16}, pages 204--222.
  Springer, 2020.

\bibitem{Sun2018c}
Deqing Sun, Xiaodong Yang, Ming-Yu Liu, and Janf Kautz.
\newblock {PWC}-{N}et: {CNN}s for optical flow using pyramid, warping, and cost
  volume.
\newblock {\em CVPR}, 2018.

\bibitem{Sundaram2010}
Narayanan Sundaram, Thomas Brox, and Kurt Keutzer.
\newblock Dense point trajectories by gpu-accelerated large displacement
  optical flow.
\newblock In {\em ECCV}, 2010.

\bibitem{teed2020raft}
Zachary Teed and Jia Deng.
\newblock {RAFT}: Recurrent all-pairs field transforms for optical flow.
\newblock In {\em European Conference on Computer Vision}, pages 402--419.
  Springer, 2020.

\bibitem{Wang2013}
Heng Wang, Alexander Kl{\"a}ser, Cordelia Schmid, and Cheng-Lin Liu.
\newblock Dense trajectories and motion boundary descriptors for action
  recognition.
\newblock {\em International journal of computer vision}, 103:60--79, 2013.

\bibitem{wang2023omnimotion}
Qianqian Wang, Yen-Yu Chang, Ruojin Cai, Zhengqi Li, Bharath Hariharan,
  Aleksander Holynski, and Noah Snavely.
\newblock Tracking everything everywhere all at once.
\newblock {\em arXiv:2306.05422}, 2023.

\bibitem{Wannenwetsch2017}
Anne~S Wannenwetsch, Margret Keuper, and Stefan Roth.
\newblock Probflow: Joint optical flow and uncertainty estimation.
\newblock In {\em Proceedings of the IEEE international conference on computer
  vision}, pages 1173--1182, 2017.

\bibitem{xu2018youtube}
Ning Xu, Linjie Yang, Yuchen Fan, Dingcheng Yue, Yuchen Liang, Jianchao Yang,
  and Thomas Huang.
\newblock Youtube-vos: A large-scale video object segmentation benchmark.
\newblock {\em arXiv preprint arXiv:1809.03327}, 2018.

\bibitem{Yang2019}
Gengshan Yang and Deva Ramanan.
\newblock Volumetric correspondence networks for optical flow.
\newblock In {\em Advances in Neural Information Processing Systems}, pages
  793--803, 2019.

\bibitem{Yang2021a}
Gengshan Yang, Deqing Sun, Varun Jampani, Daniel Vlasic, Forrester Cole, Huiwen
  Chang, Deva Ramanan, William~T Freeman, and Ce Liu.
\newblock Lasr: Learning articulated shape reconstruction from a monocular
  video.
\newblock In {\em Proceedings of the IEEE/CVF Conference on Computer Vision and
  Pattern Recognition}, pages 15980--15989, 2021.

\bibitem{Yang2022a}
Gengshan Yang, Minh Vo, Natalia Neverova, Deva Ramanan, Andrea Vedaldi, and
  Hanbyul Joo.
\newblock Banmo: Building animatable 3d neural models from many casual videos.
\newblock In {\em Proceedings of the IEEE/CVF Conference on Computer Vision and
  Pattern Recognition}, pages 2863--2873, 2022.

\bibitem{Zhang2022a}
Congxuan Zhang, Cheng Feng, Zhen Chen, Weiming Hu, and Ming Li.
\newblock Parallel multiscale context-based edge-preserving optical flow
  estimation with occlusion detection.
\newblock {\em Signal Processing: Image Communication}, 101:116560, 2022.

\bibitem{Zhao2020}
Shengyu Zhao, Yilun Sheng, Yue Dong, Eric~I Chang, Yan Xu, et~al.
\newblock Maskflownet: Asymmetric feature matching with learnable occlusion
  mask.
\newblock In {\em Proceedings of the IEEE/CVF Conference on Computer Vision and
  Pattern Recognition}, pages 6278--6287, 2020.

\end{thebibliography}
